\documentclass[runningheads]{llncs}

\usepackage{eccv}

\usepackage{eccvabbrv}

\usepackage{graphicx}

\usepackage{booktabs}
\usepackage{multirow}
\usepackage{colortbl}
\usepackage{xcolor}
\usepackage{algorithm}
\usepackage{algpseudocode}

\usepackage[accsupp]{axessibility}  %

\usepackage{hyperref}

\usepackage{orcidlink}

\def\methodname{\textsc{G3R}}

\definecolor{teal}{rgb}{0.0, 0.5, 0.5}
\definecolor{goldenrod}{rgb}{0.867, 0.769, 0.255}
\definecolor{gray}{rgb}{0.843, 0.843, 0.843}
\definecolor{brown}{rgb}{0.494, 0.259, 0.0196}
\definecolor{grey}{rgb}{0.9,0.9,0.9}

\newcommand{\goldenbullet}{\textcolor{goldenrod}{\fontsize{20}{20}\selectfont\textbullet}}
\newcommand{\graybullet}{\textcolor{gray}{\fontsize{20}{20}\selectfont\textbullet}}
\newcommand{\brownbullet}{\textcolor{brown}{\fontsize{20}{20}\selectfont\textbullet}}

\usepackage{amsmath,amsfonts,bm}

\def\eqref#1{equation~\ref{#1}}

\def\1{\bm{1}}

\def\rmI{{\mathbf{I}}}

\def\rmK{{\mathbf{K}}}

\DeclareMathAlphabet{\mathsfit}{\encodingdefault}{\sfdefault}{m}{sl}
\SetMathAlphabet{\mathsfit}{bold}{\encodingdefault}{\sfdefault}{bx}{n}

\definecolor{ForestGreen}{RGB}{34,139,34}

\begin{document}

\title{\methodname{}: Gradient Guided Generalizable Reconstruction}%

\titlerunning{G3R: Gradient Guided Generalizable Reconstruction}

\author{Yun Chen$^{1,2}\thanks{Equal contributions.}$\quad Jingkang Wang$^{1,2\star}$ \\ Ze Yang$^{1,2}$ \quad  Sivabalan Manivasagam$^{1,2}$\quad Raquel Urtasun$^{1,2}$
}

\authorrunning{Y. Chen and J. Wang et al.}

\institute{
Waabi$^{1}$ \quad University of Toronto$^{2}$ \\
\email{\{ychen, jwang, zyang, smanivasagam, urtasun\}@waabi.ai}
}
\maketitle

\begin{figure}[htbp!]
	\centering
	\vspace{-0.32in}
	\includegraphics[width=1.0\textwidth]{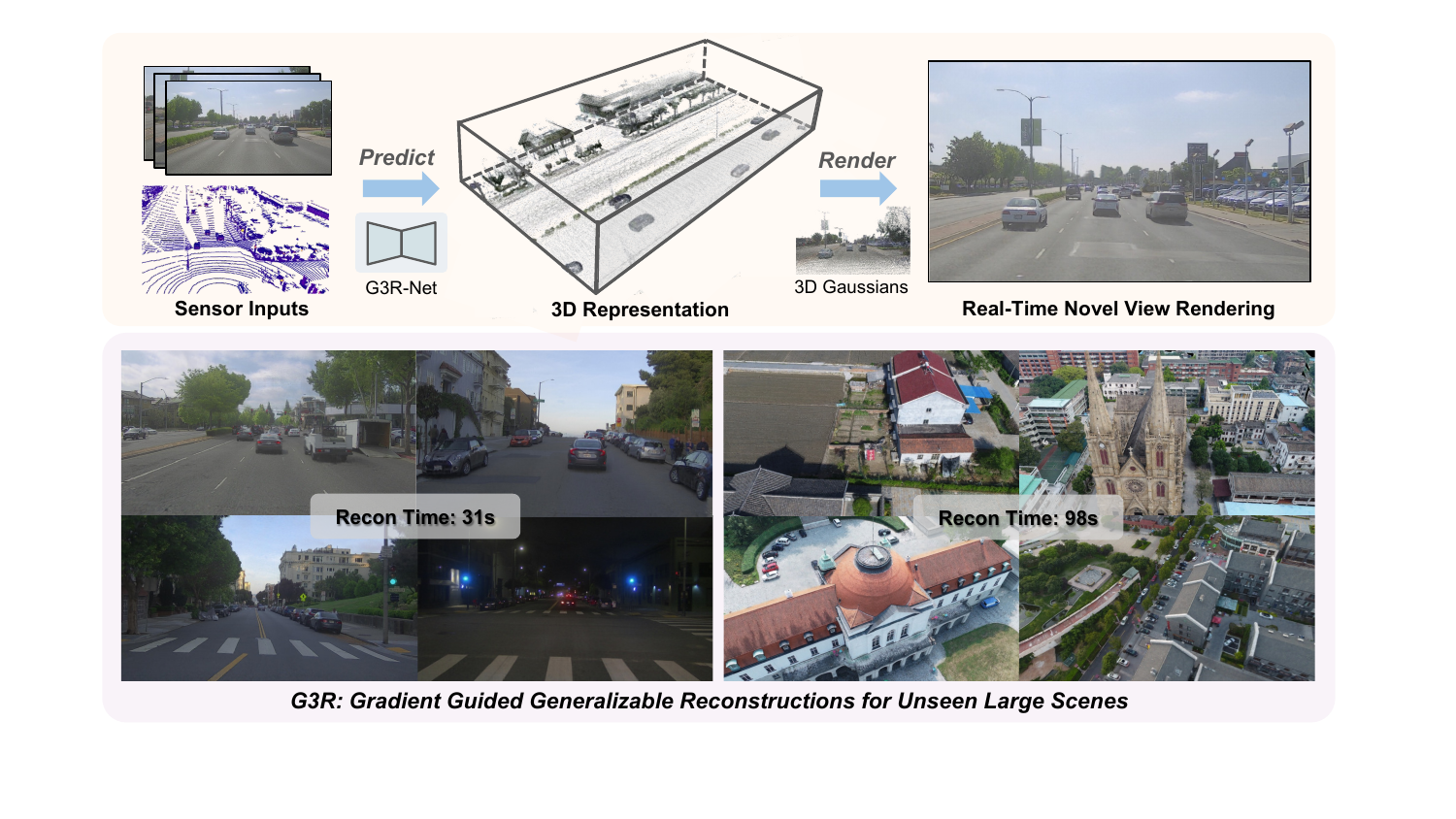}
	\vspace{-0.27in}
	\caption{\textbf{Gradient Guided Generalizable Reconstruction (\methodname{})}: Our method learns a single reconstruction network that takes multi-view camera images and an initial point set to predict the 3D representation for large scenes ($> 10,000 m^2$) in two minutes or less, enabling realistic and real-time camera simulation.
	}
	\vspace{-0.4in}
\label{fig:teaser}
\end{figure}

\begin{abstract}
Large scale 3D scene reconstruction is important for applications such as virtual reality and simulation.
Existing neural rendering approaches (\eg, NeRF, 3DGS) have achieved realistic reconstructions on large scenes, but
optimize per scene, which is expensive and slow,
and exhibit noticeable artifacts under large view changes due to overfitting.
Generalizable approaches, or large reconstruction models, are fast, but primarily work for small scenes/objects and often produce lower quality
rendering results.
In this work, we introduce \methodname{}, a generalizable reconstruction approach that can efficiently predict high-quality 3D scene representations for large scenes.
We propose to learn a reconstruction network that takes the gradient feedback signals from differentiable rendering to iteratively update a 3D scene representation, combining the benefits of high photorealism from per-scene optimization with data-driven priors from fast feed-forward prediction methods.
Experiments on urban-driving and drone datasets show that \methodname{} generalizes across diverse large scenes and accelerates the reconstruction process by at least $10 \times$
while achieving comparable or better realism compared to
3DGS, and also being more robust to large view changes. %
Please visit our project page for more results: \url{https://waabi.ai/g3r}.

\keywords{Generalizable Reconstruction \and Neural Rendering \and Learned Optimization  \and 3DGS \and Large Reconstruction Models}
\end{abstract}

\section{Introduction}
\label{sec:intro}

Reconstruction of large real world scenes from sensor data, such as urban traffic scenarios,
 is a long-standing problem in computer vision and computer graphics. %
Scene reconstruction enables applications such as virtual reality
and high-fidelity camera simulation, where robots such as autonomous vehicles can learn and be evaluated safely at scale~\cite{wang2021advsim,manivasagam2023towards,xiong2023learning,sarva2023adv3d,ljungbergh2024neuroncap}.
To be effective, the 3D reconstructions must have high photorealism at novel views,
be efficient to generate, enable scene manipulation,
and enable real-time image rendering.

Recently, neural rendering approaches such as NeRF\cite{mildenhall2020nerf} and 3D Gaussian Splatting (3DGS)\cite{3dgs} have achieved realistic reconstructions for large scenes using camera and optionally LiDAR data.
However, they require a costly per-scene optimization process to reconstruct the scene by recreating the input sensor data via differentiable rendering, which may take several hours
to achieve high-quality. Moreover, they typically focus on the novel view synthesis (NVS) setting where the target view is close to the source views and often exhibit artifacts when the viewpoint changes are large (\eg, meter-scale shifts), as it can overfit to the input images while not learning the true underlying 3D representation.

To enable faster reconstruction and better performance at novel views,
recent works aim to synthesize a \textit{generalizable} representation with a single pre-trained network, which can be used for NVS
on unseen scenes in a zero-shot manner.
These methods utilize an encoder to predict the intermediate scene representation
 by aggregating image features extracted from multiple source views according to camera and geometry priors, and then decode the representation for
  NVS via volume rendering or a transformer
 ~\cite{chen2021mvsnerf,yu2021pixelnerf,wang2021ibrnet,lin2022efficient,wang2022attention}.
The encoder and decoder networks are trained across many %
scenes to  reconstruction priors.
Most recently, large reconstruction models (LRMs) are proposed to learn reconstruction priors by training on large-scale synthetic datasets for generalizable single-step 2D to 3D reconstruction \cite{liu2023zero, hong2024lrm,li2024instantd,zhang2024gs, wei2024meshlrm}.
However, both  generalizable NVS and LRMs are primarily
applied to objects or small scenes due to the complexity of
large scenes,
which are difficult to predict accurately from a single step network prediction.
Furthermore, the computation resources and memory
needed to
utilize many input scene images (>100)
with existing techniques that aggregate ray features \cite{wang2022attention}, build cost volumes \cite{chen2021mvsnerf} or perform image-based rendering \cite{wang2021ibrnet} are prohibitive. %

In this paper, we present Gradient Guided Generalizable Reconstruction (\methodname{}), the first method that enables fast and generalizable reconstruction of large scenes.
Given a sequence of images and an approximate geometry scaffold (\eg, points from LiDAR or multi-view stereo), \methodname{} can produce a modifiable digital twin as a set of 3D Gaussian primitives in two minutes or less for large scenes ($>10,000m^2$).
This representation can
be directly used for high-fidelity novel-view rendering at interactive frame rates ($>90$ FPS). %
Our key idea is to learn a single reconstruction network
that iteratively updates the 3D scene representation,
combining the benefits of data-driven priors from fast
prediction methods with the iterative gradient feedback signal
from per-scene optimization methods.
\methodname{} can be viewed as a ``learned optimizer''~\cite{wichrowska2017learned,andrychowicz2016learning} for scene reconstruction.
Towards this goal, we first initialize a
neural scene representation which we call
\textit{3D Neural Gaussians} from the geometry scaffold that can be differentiably rendered. %
Rather than select a few close-by source views for unprojection like existing generalizable works, we propose a novel way of lifting 2D images to 3D space by rendering and backpropagating to obtain gradients w.r.t the current 3D representation.
These 3D gradients can be seen as 2D images unprojected to 3D with the current representation as the 3D proxy, which takes the rendering procedure into account, and is thus naturally occlusion aware and contains a useful feedback signal.
Moreover, it provides a unified representation that can efficiently aggregate as many 2D images as needed by just aggregating the gradients.
Then, our reconstruction network (\methodname-Net) takes the 3D gradients and current 3D representation as inputs and iteratively predicts updates to refine the representation.
Since the \methodname-Net incorporates the rendering feedback signal at each step and is trained across multiple scenes, it can significantly
 accelerate the convergence compared to standard gradient descent algorithms (\textit{i.e}, 24 iterations v.s. 1000s of iterations).
\methodname-Net is trained across multiple scenes,
enabling high quality reconstruction and
improving robustness for NVS.

Experiments on two outdoor datasets with large-scale scenes demonstrate the generalizability of \methodname{}.
With as little as 24 iterations,
\methodname{} reconstructs large scenes with comparable or better realism at novel views than the per-scene optimization approaches while being  at least $10\times$ faster.
To the best of our knowledge, this is the first generalizable reconstruction approach
that can reconstruct a faithful 3D
representation for such large-scale scenes ($>10,000 m^2$) in high-resolution ($>100$ source images at $1080\times1920$),
showing the potential to
build digital twins for the metaverse and simulation at large scale.

\begin{figure}[tb]
	\centering
	\includegraphics[width=0.99\textwidth]{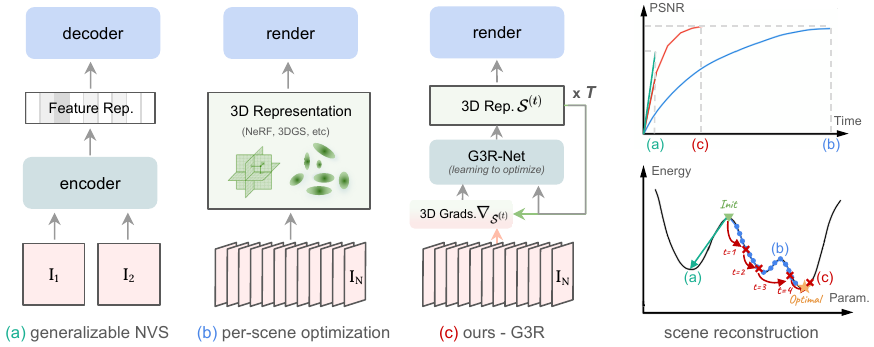}
	\vspace{-0.1in}
	\caption{\textbf{Three paradigms for scene reconstruction and novel view synthesis (NVS).} (a) Existing generalizable approaches select a few reference images (usually $\le 5$) for feed-forward prediction of intermediate representation and then decode/render the feature representation to produce the rendered images. (b) Per-scene optimization approaches take all source images (\eg, $> 100$ for large scenes) and reconstructs a 3D representation via energy minimization and differentiable rendering. (c) \methodname{} conducts iterative prediction to refine the 3D representation with the 3D gradient guidance (\ie, learned optimization) taking all source images. Compared to the other two paradigms, \methodname{} leverages the benefits of both worlds (data-driven priors, gradient feedback) and achieves the best trade-off between the reconstruction quality and time (rightmost).
	}
	\vspace{-0.1in}
	\label{fig:pipeline_comp}
\end{figure}

\section{Related Work}
\label{sec:related}

\paragraph{Optimization-based scene reconstruction:}
The current state-of-the-art in scene reconstruction is optimizing differentiable radiance fields,
such as NeRF~\cite{mildenhall2020nerf} or 3DGS~\cite{3dgs}, which model the 3D scene either as neural networks or as Gaussian primitives, and then alpha-composite along the ray via either ray-marching or rasterization, respectively.
To extend to city-scale
scenes, some works
decompose the scene into sub-components and represent each with a network
to increase model capacity~\cite{tancik2022block,turki2022mega,lin2024vastgaussian,zhenxing2022switch}.
To enable realistic and controllable sensor simulation, another line of work
decomposes dynamic scenes (\eg, urban driving scenes) into static background and moving objects~\cite{ost2021neural,unisim,wu2023mars,tonderski2023neurad,huang2023neural,zhou2023drivinggaussian,yang2023emernerf,liu2023real,yan2024street,yang2023reconstructing,yang2021recovering} or conduct inverse rendering for geometry, material, lighting and semantics decomposition~\cite{wang2022cadsim,yang2023reconstructing,wang2023neural,pun2023neural,lin2023urbanir}.
These works require time-consuming (hours or days) per-scene optimization for large scenes and often exhibit artifacts at large view changes due to overfitting.
In contrast, \methodname{}
predicts a high-quality and robust 3D representation for large scenes in a few minutes or less.

\paragraph{Generalizable reconstruction:}
To generalize to novel scenes, researchers train neural networks across diverse scenes and incorporate proxy geometry like depth maps for image-based rendering~\cite{riegler2020fvs,riegler2021svs,wiles2019synsin, aliev2020neural, kopanas2021pointbased}.
However, it is usually challenging or expensive to obtain high-quality geometry for real-world large scenes.
To address this issue, recent works adopt transformers to either directly map the source images and camera embedding to the target view without any physical constraints~\cite{srt22, sajjadi2022rust,  rombach2021geometryfree,kulhanek2022viewformer, seitzer2023dyst} or aggregate points from source images along the epipolar lines for rendering~\cite{wang2021ibrnet, suhail2022generalizable, suhail2021light, wang2022attention, cong2023enhancing,sitzmann2021light,niemeyer2020differentiable,yang2023contranerf,trevithick2021grf,reizenstein2021common, chibane2021stereo}.
Another popular approach is to lift 2D images to 3D cost volumes with geometry priors~\cite{chen2021mvsnerf, lin2022efficient, chibane2021stereo, johari2022geonerf, liu2022neural} but struggles with large camera movement. These
methods do not produce a unified 3D representation, suffer from noticeable artifacts under large view changes, and are slow to render.
On the other hand, some works that directly predict 3D representations such as multi-plane images (MPI)~\cite{zhou2018stereo,srinivasan2019pushing,flynn2019deepview} or implicit representations~\cite{SitzmannZW19SRN,niemeyer2020differentiable,mller2022autorf, chen2021mvsnerf, yang2021s3} only work well on objects or small scenes.
Concurrent work \cite{charatan2023pixelsplat} predicts 3D Gaussians for generalizable reconstruction, but is limited to low-resoluation image pairs.
In contrast, \methodname{} take all available source images and predicts a unified representation for large-scale scenes including dynamics, enabling scalable and realistic simulation.
Most recently, large reconstruction models~\cite{liu2023zero, hong2024lrm,li2024instantd,zhang2024gs, wei2024meshlrm} (LRMs) achieve strong generalizability across small objects by training on large synthetic dataset such as Objaverse. To our best knowledge, G3R is the first LRM that generalizes across diverse large scenes and handles large view changes by training on large-scale real-world datasets.

\paragraph{Iterative networks for 3D:}

Our method falls under the ``iterative network'' framework, which conduct iterative updates to gradually refine the output.
Prior works have studied iterative approaches on low-dimensional inverse problems~\cite{carreira2015human,adler2017solving, manhardt2018deep,li2018deepim,ma2021deep} such 6-DOF pose and illumination estimation. %
In contrast, \methodname{} solves a challenging high-dimensional inverse problem (\ie, scene reconstruction) using a learned optimizer~\cite{andrychowicz2016learning,li2016learning,wichrowska2017learned}.
Specifically, we train a neural network that exploits spatial correlation to expedite the
reconstruction process.
Similar to \methodname, DeepView \cite{flynn2019deepview}
also employs an iterative network with gradient guidance to reconstruct a 3D representation (MPI), but for small baselines only.
Moreover, it
unfolds the optimization
through a series of distinct CNN networks and loss-agnostic gradient components at each stage for each source image, limiting the number of input images, and leading
to large memory usage and slow speed.

\section{Gradient Guided Generalizable Reconstruction (\methodname{})}

\begin{figure}[tb]
	\centering
	\includegraphics[width=0.99\textwidth]{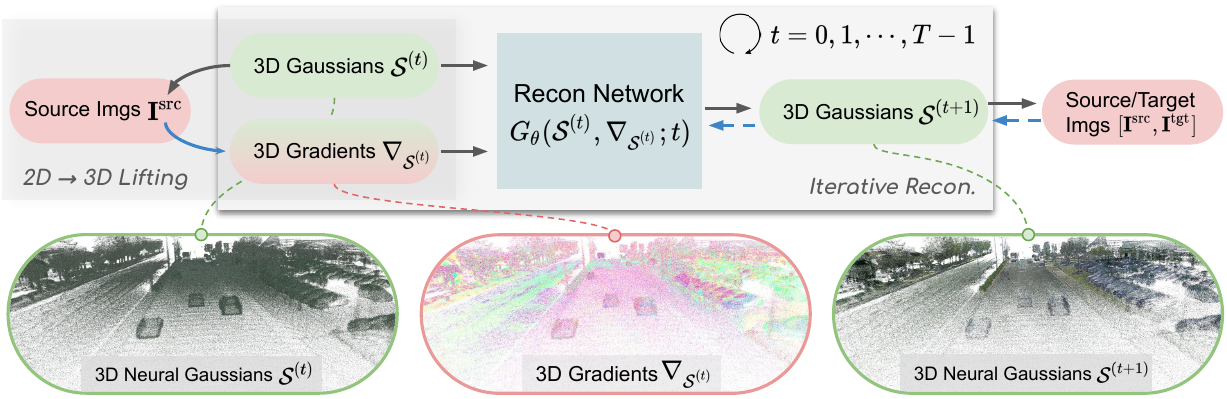}
	\vspace{-0.07in}
	\caption{\textbf{Method overview.}
		We model the generalizable reconstruction as an iterative process,
		where the 3D neural Gaussians $\mathcal S^{(t)}$ are iteratively refined with reconstruction network $G_\theta$.
		We first lift the source 2D images $\rmI^{\mathrm{src}}$ to 3D space by backpropogating the rendering procedure to get the gradients w.r.t the representation $\nabla_{\mathcal S^{(t)}}$ (\textcolor{blue}{blue arrow}).
		Then the reconstruction network $G_\theta$ takes the 3D representation $\mathcal S^{(t)}$, the gradient $\nabla_{\mathcal S^{(t)}}$ and the iteration step $t$ as input, and predicts an updated 3D representation $\mathcal{S}^{(t+1)}$.
		To train the network, we render $\mathcal{S}^{(t+1)}$ at source and novel views, and compute loss. The backward gradient flow for training $G_\theta$ is highlighted with \textcolor{blue}{dashed blue arrows}.
	}
	\label{fig:method}
	\vspace{-0.15in}
\end{figure}

Given a set of source
camera images $\rmI^{\text{src}} = \{\rmI_i\}_{1\leq i \leq N}$ and an approximate geometry scaffold $\mathcal{M}$ (\eg, obtained from either LiDAR
or points from multi-view stereo)
captured in-the-wild  by a sensor platform moving through a large dynamic scene,
our goal is to efficiently reconstruct
a realistic and editable 3D representation $\mathcal{S}$
for accurate real-time camera simulation.
In this paper, we introduce Gradient Guided Generalizable Reconstruction (\methodname{}),  %
the first method that can create modifiable digital clones of large real world scenes ($>10,000m^2$) in two minutes or less, and that renders novel views with high photorealism at >90 FPS.
Our method overview is shown in \cref{fig:method}.
\methodname{} combines
data-driven priors from fast
prediction methods with the iterative gradient feedback signal from per-scene optimization methods by learning to optimize for large scene reconstruction (\cref{fig:pipeline_comp}-left).
\methodname{}
iteratively updates a representation we call 3D
neural Gaussians,
initialized from the scaffold $\mathcal M$, with a single neural network.
The network takes the gradient feedback signals from differentiably rendering the representation to reconstruct the source images $\rmI^{\text{src}}$.
\methodname{}  achieves the best trade-off between realism and reconstruction speed, achieving performance and scalability (see \cref{fig:pipeline_comp}-right). %

In what follows, we first introduce our
scene representation (3D neural Gaussians)  designed for handling dynamic and unbounded large scenes
(\cref{sec:scene_rep}).
Then we show how to lift 2D images to 3D space by propagating the gradients  (\cref{sec:gradient_lift}), followed by iterative refinements in \cref{sec:iterative_recon}.
We describe training the network across multiple scenes
in \cref{sec:training}.

\subsection{\methodname{}'s Scene Representation}
3D Gaussian Splatting~\cite{3dgs} (3DGS) is a differentiable rasterization technique that allows real-time rendering of photorealistic scenes learned from posed images and an intitial set of points from
SfM~\cite{snavely2006phototourism}.
3DGS represents the scene {with a set of 3D Gaussians (\ie, points) $\mathcal{G} = \{g_i\}_{1 \leq i \leq M}$,
where $g_i \in \mathbb R^{14}$ consists of position ($\mathbb R^3$), scale ($\mathbb R^3$), orientation ($\mathbb R^4$), color ($\mathbb R^3$) and opacity ($\mathbb R^1$).
}
These gaussian points $\mathcal{G}$ can be rendered to 2D images with camera poses $\mathrm{\Pi}$ using a differentiable tile rasterizer $f_{\mathrm {rast}} (\mathcal{G}, \mathrm{\Pi} )$, where each point is projected and splatted to the image plane based on the scale and orientation, then the color is blended with other points based on the opacity and depth to camera.
However, 3DGS's explicit representation lacks modelling capacity useful for learning-based optimization.
Furthermore, 3DGS~\cite{3dgs} focuses on
small static scenes or individual  objects, and has challenges
modeling large-scale dynamic scenes,
 such as self-driving scenarios.
 In this paper, we make two enhancements to 3DGS's representation.
 First, we augment its representation with a latent feature vector, which we call \emph{3D neural Gaussians}, providing additional
 capacity for generalizable reconstruction and learning-based optimization.
 Second, we decompose the scene into the nearby static scene, dynamic actors, and a distant region to enable modelling of large unbounded dynamic scenes.
 We now describe these two enhancements and then detail the rendering process.%

\paragraph{3D Neural Gaussians:}
\label{sec:scene_rep}
We define our scene representation $\mathcal{S}$ as
a set of \textit{3D Neural Gaussians, }
$\mathcal{S} = \{h_i\}_{1 \leq i \leq M}$, where each point is represented by a feature vector $h_i \in \mathbb R^{C}$.
This latent representation helps encode information about the scene during the iterative updates in the learning-based optimization described in Sec.~\ref{sec:iterative_recon}.
To render, we convert the 3D neural Gaussians to a set of explicit color 3D Gaussians $\mathcal{G} = \{g_i\}_{1 \leq i \leq M}$,
using a Multi-Layer Perceptron (MLP) network
$g_i = f_\mathrm{mlp}(h_i)$.
To encode geometry and additional physical information about the scene into $h_i$ and ensure stable optimization, we designate the first 14 channels as the 3D Gaussian attributes and add a skip connection in $f_\mathrm{mlp}$ such that it updates these channels to generate $g_i$.

\paragraph{Representing rigid dynamic objects and unbounded scenes:}
We decompose the dynamic scene and its set of 3D neural Gaussians $\mathcal{S}$ into a static background $\mathcal{S}^\mathcal{B}$,
a set of dynamic actors $\mathcal{S}^\mathcal{A}$ and a distant region $\mathcal{S}^\mathcal{Y}$ (\eg, far-away buildings and sky).
We assume rigid motion $\mathcal{T} (\mathcal{S}^\mathcal{A}, \boldsymbol \xi^{\mathcal{A}}) $ for dynamic actors, where $\mathcal{T}$ is the rigid transformation and $\boldsymbol \xi^{\mathcal{A}}$ are the actor extrinsics.
The dynamic points $\mathcal{S}^\mathcal{A}$ are moved
across different frames using
3D bounding boxes that specify each foreground actor's size and location.
We initialize the 3D neural Gaussians
for the static background and dynamic actors using the provided approximate geometry scaffolds $\mathcal M$ (\eg, aggregated LiDAR points or multi-view stereo points).
We further position a fixed number of points at a large distance to model the distant region.
See
Sec.~\ref{sec:exp} and Appendix~\ref{sec: g3r_supp_implement} for details.

\paragraph{Rendering:}
Given
$\mathcal{S}$ and camera poses $\mathrm{\Pi} = \{\rmK_i, \boldsymbol{\xi}_i\}$, where $\rmK_i$ and $\boldsymbol{\xi}_i$ are the camera intrinsics and extrinsics for view $i$, we
convert $\mathcal S$ to 3D Gaussians $\mathcal{G}$ and then leverage
the differentiable tile rasterizer \cite{3dgs} to render the images $\hat{\rmI}$: %
\begin{align}
f_{\mathrm{render}}(\mathcal{S}; \mathrm{\Pi}) := &f_{\mathrm{rast}}(\mathcal G; \mathrm{\Pi}) = f_{\mathrm{rast}}(f_{\mathrm{mlp}} (\mathcal{S}); \mathrm{\Pi}) \\
=& f_{\mathrm{rast}}(f_{\mathrm{mlp}} (\mathcal{S}^{\mathcal{B}}, \mathcal{S}^{\mathcal{Y}}, \mathcal{T}(\mathcal{S}^\mathcal{A}, \boldsymbol \xi^{\mathcal{A}})); \mathrm{\Pi})
\label{eqn:render_eqn}
\end{align}

\subsection{Lift 2D Images to 3D as Gradients}
\label{sec:gradient_lift}

Previous generalizable works~\cite{chen2021mvsnerf,lin2022efficient,wang2022attention} lift a few 2D images (\eg, $\leq$ 5) to 3D by aggregating image features extracted from source views according to camera and geometry priors (\eg, epipolar geometry or multi-view stereo).
Since each image is processed by a neural network separately, it cannot take many source images due to the high memory usage in both training and inference, limiting its applicability to
small objects under small viewpoint changes. %
This is because large scenes usually have complex topology/geometry and cannot be reconstructed accurately with only a small set of source images.
Moreover,
it can be
challenging to select and merge source views
and also ensure spatial consistency.

Instead, we propose to lift 2D images to 3D space by ``rendering and backpropagating'' to obtain gradients w.r.t the 3D representation.
Compared to leveraging networks to process images independently, 3D gradients provide a unified representation
that can efficiently aggregate as many images as needed.
Moreover, 3D gradients take the rendering procedure into account, naturally handling occlusions.
It also enables adjustment of the 3D representation, which is not done in
traditional depth rendering for view warping.
Finally, the 3D gradients are fast to compute with modern differentiable rasterization engines.

Specifically, given the 3D representation $\mathcal{S}$, we first render the scene to source input views $\hat{\rmI}^{\mathrm{src}} = f_{\mathrm{render}}(\mathcal{S}; \mathrm{\Pi}^{\mathrm{src}})$ using Eqn.~\ref{eqn:render_eqn}.
Then, we compare the rendered images with the inputs $\rmI^{\mathrm{src}}$, compute the reconstruction loss $L$, and backpropagate the difference to 3D representation $\mathcal{S}$ to get accumulated gradients $\nabla_{\mathcal{S}} := \nabla_{\mathcal{S}} L(\mathcal{S}, \rmI^{\mathrm{src}}; \mathrm{\Pi}^{\mathrm{src}})$ as shown in \cref{fig:method}, with
\begin{align}
\label{equ:lift}
L(\mathcal{S}, \rmI^{\mathrm{src}}; \mathrm{\Pi}^{\mathrm{src}}) &=  \sum_i \left\lVert \rmI_i^{\mathrm{src}} - \hat{\rmI}_i^{\mathrm{src}} \right\rVert_2 = \sum_i \left\lVert \rmI_i^{\mathrm{src}} - f_{\mathrm{render}}(\mathcal
S; \mathrm{\Pi}_i^{\mathrm{src}}) \right\rVert_2, \\
\nabla_{\mathcal{S}} L(\mathcal S, \rmI^{\mathrm{src}}; \mathrm{\Pi}^{\mathrm{src}}) &= \frac{\partial L(\mathcal S, \rmI^{\mathrm{src}}; \mathrm{\Pi}_i^{\mathrm{src}}) }{\partial \mathcal S}= \sum_i \frac { \partial\left\lVert\rmI_i^{\mathrm{src}} -  f_{\mathrm{render}}(\mathcal S; \mathrm{\Pi}_i^{\mathrm{src}})\right\rVert_2} {\partial \mathcal{S}}.
\end{align}
The differentiable function  $f_{\mathrm{render}}$ builds a connection between 2D and 3D, and the gradient $\nabla_\mathcal{S}$
encodes the 2D images in 3D
using $\mathcal{S}$ as the proxy.

\subsection{Iterative Reconstruction with a Neural Network}
\label{sec:iterative_recon}
We now describe how we iteratively refine the scene representation $\mathcal{S}$ given the source images $\rmI^{\mathrm{src}}$.
At each step $t$, we take the current 3D representation $\mathcal{S}^{(t)}$  as a proxy to compute the gradient $\nabla_{{\mathcal{S}}^{(t)}}$ via differentiable rendering, thereby unprojecting 2D
source images $\rmI^{\mathrm{src}}$ to 3D, and then feed $\nabla_{{\mathcal{S}}^{(t)}}$ into the network $G_\theta$ to predict the updated
 3D representation $\mathcal{S}^{(t+1)}$:
\begin{equation}
\mathcal S^{(t+1)}  = \mathcal S^{(t)} + \gamma(t) \cdot G_\theta(\mathcal S^{(t)},
\nabla_{\mathcal{S}^{(t)}} L(\mathcal{S}^{(t)}, \rmI^{\mathrm{src}}; \mathrm{\Pi}^{\mathrm{src}})
; t), \  \ t = 0, 1, \dots, T-1.
\end{equation}
$\gamma(t)$ defines the update scale at different step $t$. Intuitively, similar to gradient descent, we desire a decaying schedule $\gamma(t)$ and a small $T$ so that the network can predict an initial coarse representation and then quickly refine it.
We use the cosine scheduler from DDIM~\cite{song2020denoising} for $\gamma(t)$.
We use a 3D UNet\cite{cciccek20163d} with sparse convolution~\cite{tangandyang2023torchsparse} as $G_\theta$ to process the neural Gaussians $\mathcal{S}$.
The iterative process allows us to refine the 3D representation to achieve better quality and use a smaller network that is more efficient and easier to learn.

\subsection{Training \& Inference}
\label{sec:training}
We now describe the training process to train the learned optimizer $G_\theta$ and neural decoding MLP $f_\text{mlp}$.
For each scene, we initialize the scene representation $\mathcal{S}^{(0)}$ from the geometry scaffold $\mathcal M$.
We iteratively refine $\mathcal{S}$ with the network prediction for $T$ steps.
To enhance the generalizability of reconstruction network, we render the updated representation to both source views $\rmI^{\mathrm{src}}$ and novel views $\rmI^{\mathrm{tgt}}$ during training $(\rmI = \left[\rmI^{\mathrm{src}}, \rmI^{\mathrm{tgt}}\right])$, and backpropagate
the gradients to the parameters of the reconstruction network $G_\theta$ and the
	$f_\mathrm{mlp}$.
Note that in Eqn.~\ref{equ:lift} only the gradients from source views are used as input to $G_\theta$ for the next iteration, as the target views will not be available at test time.
$G_\theta$ is trained to minimize final rendering loss for every iteration step $t$.
We train the networks across many large outdoor scenes.
The total loss $\mathcal{L}$ is:
\begin{equation}
\label{equ:loss}
\mathcal{L} = \mathcal{L}_\mathrm{mse}(\hat \rmI, \rmI) + \lambda_{\mathrm{lpips}} \mathcal{L}_\mathrm{lpips} (\hat \rmI, \rmI) + \lambda_{\mathrm{reg}} \mathcal{L}_\mathrm{reg} (\mathbf{\mathcal{G}}),
\end{equation}
where $\hat \rmI$ is the rendered images, $\mathcal{L}_\mathrm{mse}$ is the photometric loss, $\mathcal{L}_\mathrm{lpips}$ is the perceptual loss\cite{zhang2018unreasonable}, and $\mathcal{L}_\mathrm{reg}$ is the regularization term applied on the shape of the transformed Gaussians $\mathcal{G}$ to be flat for better alignment with the surface.
\begin{equation}
	\mathcal{L}_\mathrm{reg} (\mathcal{G}) =  \sum_i \max(0, d_i^{\min} - \epsilon),
\end{equation}
where $d_i^{\min}$ is the minimal value of the 3-channel scale for each Gaussian $g_i$. We encourage it to be smaller than a threshold $\epsilon$.

\paragraph{Inference:} %

Given the pre-trained reconstruction network $G_\theta$ and neural Gaussian decoder
MLP $f_\mathrm{mlp}$, we can now reconstruct novel scenes not seen during training.
Specifically, we take all input images $\rmI^{\mathrm{src}}$ for the novel scene and the 3D neural Gaussian initialization $\mathcal{S}^{(0)}$ to iteratively compute the gradients $\nabla_{{\mathcal S}}$ and refine the 3D representation.
Finally, we export $\mathcal S^{(T)}$ to standard 3D Gaussians $\mathcal{G}^{(T)}$ for real-time rerasterization.

\section{Experiments}
\label{sec:exp}

We compare \methodname{} against state-of-the-art (SoTA) generalizable and per-scene optimization approaches, ablate our design choices, and demonstrate the capability of generalization across datasets.
Finally, we show that \methodname{}-predicted representation is editable and we can generate realistic multi-camera videos. %

 \begin{table}[t]
	\caption{\textbf{Comparison to reconstruction methods on PandaSet.} The methods with best photorealism are marked using gold\raisebox{-0.7ex}{\goldenbullet}, silver\raisebox{-0.7ex}{\graybullet}, and bronze\raisebox{-0.7ex}{\brownbullet} medals. $\dagger$ denotes the method needs to reconstruct the scene again with different source images when rendering each new view.
}
\vspace{-0.13in}
	\resizebox{\textwidth}{!}{
		\setlength{\tabcolsep}{4pt}
		{\renewcommand{\arraystretch}{1.05}
			\begin{tabular}{@{}ll|lll|cc@{}}
				\toprule
				\multicolumn{2}{l|}{\multirow{2}{*}{Models}}   & \multicolumn{3}{c|}{Novel View Synthesis} & \multicolumn{2}{c}{Inference Time} \\
				&                 & PSNR$\uparrow$      & SSIM$\uparrow$     & LPIPS$\downarrow$     & Recon Time           & Render FPS      \\ \midrule
				\multicolumn{1}{l|}{\multirow{4}{*}{Generalizable}}      & MVSNeRF$_{ft}$~\cite{chen2021mvsnerf}        &   23.68  &  0.659  &  0.482       &  35min 31s &  0.0392           \\
				\multicolumn{1}{l|}{} & ENeRF~\cite{lin2022efficient}           &   24.43   &  0.736\raisebox{-0.6ex}{\brownbullet}   &  0.306\raisebox{-0.6ex}{\goldenbullet}   &   0.057s$^\dagger$     &    6.93          \\
				\multicolumn{1}{l|}{} & GNT~\cite{wang2022attention}              &           23.99 & 0.693 &	0.408  &  0.32s$^\dagger$      & 0.00498 \\
				\multicolumn{1}{l|}{} & PixelSplat~\cite{charatan2023pixelsplat}      &  23.21     & 0.653  &  0.490  &   0.74s$^\dagger$     &  147   \\ \midrule
				\multicolumn{1}{l|}{\multirow{2}{*}{Per-scene Opt.}} & Instant-NGP~\cite{mueller2022instant}    & 24.34  &  0.729      & 0.436 &  7min 16s   &  3.24 \\
				\multicolumn{1}{l|}{} & 3DGS~\cite{3dgs}     &  25.14\raisebox{-0.6ex}{\graybullet}  &  0.747\raisebox{-0.6ex}{\goldenbullet}   &  0.372\raisebox{-0.6ex}{\brownbullet}     & 50min 14s &  121               \\ \midrule
				\multicolumn{1}{l|}{\multirow{2}{*}{Ours}}  & \methodname{} (turbo) & 24.76\raisebox{-0.6ex}{\brownbullet} &	0.720 &	0.438   &   31s &  121 \\
				\multicolumn{1}{l|}{}  &  \methodname{}  &  25.22\raisebox{-0.6ex}{\goldenbullet} &   0.742\raisebox{-0.6ex}{\graybullet}  & 0.371\raisebox{-0.6ex}{\graybullet} &   123s   &     121      \\ \bottomrule
			\end{tabular}
		}
	}
	\vspace{0.02in}
	\label{tab:pandaset_comp}
\end{table}

\subsection{Experimental Setup}
\paragraph{Datasets:}
We conduct experiments on two public datasets with large real-world scenes:
PandaSet~\cite{xiao2021pandaset}, which contains dynamic actors in driving scenes and BlendedMVS~\cite{yao2019blendedmvs}, which contains large static infrastructure.
We select 7 diverse scenes
for testing with each  covering around $200\times 80m^2$,
and the rest (96 scenes) for training.
BlendedMVS-large is a collection of 29 real-world scenes captured by a drone, ranging in size from $10,000 m ^2$ to over $100,000 m^2$, and also includes
reconstructed meshes from multi-view stereo~\cite{altizure}.
We select 25 scenes for training and 4 for testing.
For both datasets, we use every other frame as source and the remaining for test.
BlendedMVS has more challenging novel views, as the distance
between two nearby views can be large (See Appendix~\ref{fig:blendedmvs}). %

\paragraph{Implementation details:}
We initialize the 3D neural Gaussians' $\mathcal{S}^{(0)}$
positions using downsampled 3D points from LiDAR points in PandaSet or mesh faces in BlendedMVS.
To ensure geometry coverage,
the scale for each Gaussian is initialized isotropically
as the distance to its third nearest point.
The rotation is set to identity and the opacity to 0.7.
The other feature channels are randomly initialized.
We disable view-dependent spherical harmonics from the original 3DGS \cite{3dgs} for simplicity and improved memory usage.
We normalize the 3D gradients $\nabla_{\mathcal{S}^{(t)}} L(\mathcal{S}^{(t)})$ by channel across all the points before feeding to the network.
For dynamic scenes, we adopt 3 separate networks for the background, actors, and the distant region. %
\texttt{tanh} activation is applied in the output layer.
The per-scene reconstruction step $T$ is set as $24$ during training.
We train for 1000 scene iterations in total using Adam optimizer \cite{kingma2014adam} with learning rate 1e-4.
This takes roughly 30 hours on 2 RTX 3090 GPUs.
We
adopt a warm-up strategy during training that gradually increases the scene reconstruction steps
in the first few scene iterations.
The network is updated at each reconstruction step.
We provide two variants during evaluation, where the faster model, \methodname{}~(turbo), uses fewer iterations and fewer
3D neural Gaussians.
See Appendix~\ref{sec: g3r_supp_implement}.

\begin{figure}[t]
	\centering
	\includegraphics[width=0.99\textwidth]{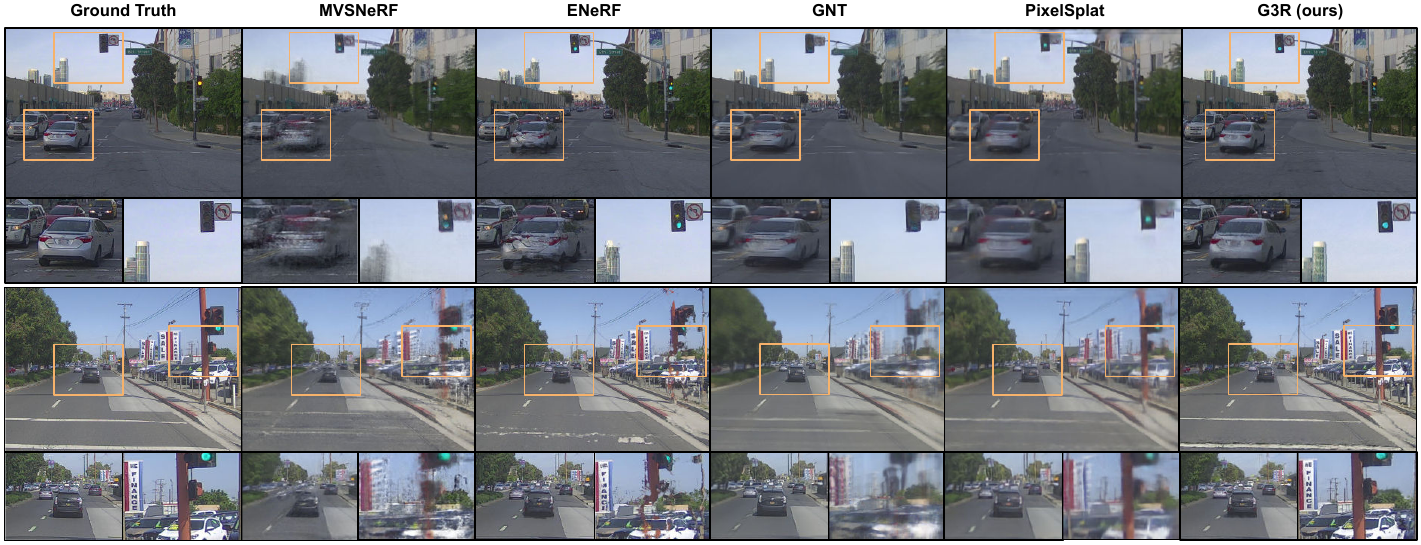}
	\vspace{-0.1in}
	\caption{\textbf{Qualitative comparison to generalizable approaches on PandaSet.}}
	\vspace{-0.15in}
	\label{fig:pandaset_comp}
\end{figure}

\paragraph{Baselines:} We compare \methodname{}
against both generalizable NVS (\cref{fig:pipeline_comp}a) and per-scene optimization approaches (\cref{fig:pipeline_comp}b).
For generalizable NVS, we compare against MVSNeRF~\cite{chen2021mvsnerf}, ENeRF~\cite{lin2022efficient}, GNT~\cite{wang2022attention} and
concurrent work PixelSplat~\cite{charatan2023pixelsplat}.
MVSNeRF warps 2D image features onto a plane sweep and then applies a 3D CNN to reconstruct a NeRF which can be finetuned further. %
Similarly, ENeRF also warps multi-view source images and leverages
depth-guided sampling for efficient reconstruction and rendering.
GNT samples points along each target ray and predicts the pixel color by learning the aggregation of view-wise features from the epipolar lines using transformers.
PixelSplat predicts 3D Gaussians with
 a 2-view epipolar transformer to extract features and then predict the depth distribution and pixel-aligned Gaussians.
Except for MVSNeRF, which finetunes the predicted representation on new scenes,
all generalizable methods
need to reconstruct the scene again with different nearest neighboring source images when rendering a new view.
Unless stated otherwise, we train and evaluate all generalizable models using the same data as \methodname{}.
For per-scene optimization approaches, we compare against
Instant-NGP~\cite{mueller2022instant} and 3DGS~\cite{3dgs}.
Instant-NGP is an
efficient NeRF framework with multi-hash grid encoding and tiny MLP for fast reconstruction.
We enhance Instant-NGP with depth supervision for better performance.
3DGS models the scene with 3D Gaussians and uses a differentiable rasterizer for fast scene reconstruction and real-time rendering.
We enhance 3DGS to support dynamic actors and unbounded scenes with the same implementation as \methodname{}.
We optimize each test scene separately using all source frames.
Please see Appendix~\ref{sec:baselines} for additional details.

\subsection{Generalizable Reconstruction on Large Scenes}

\paragraph{Scene Reconstruction on PandaSet:}

We report scene reconstruction results on PandaSet in \cref{tab:pandaset_comp} and \cref{fig:pandaset_comp}.
Compared to SoTA generalizable approaches, \methodname{} achieves significantly better photorealism and real-time rendering with an affordable reconstruction cost ($2$ min or less).
In contrast, baselines conduct image-based rendering and result in noticeable artifacts for dynamic actors due to the lack of explicit 3D representation that can model dynamics.
Moreover, they often produce blurry rendering results, especially in nearby regions where there are large view changes, due to flawed representation prediction
and poor geometry estimation for view warping.
We note that ENeRF achieves good LPIPS with image warping, but has severe visual artifacts and low PSNR.
We also compare \methodname{} with SoTA per-scene optimization approaches including Instant-NGP and 3DGS.
Our approach
achieves
on par or better photorealism while shortening the reconstruciton time to 2 minutes.
We note that PixelSplat leads to a higher FPS since it can only process low-resolution images and predicts a smaller number of 3D Gaussian points compared to \methodname{} due to memory limitations.

\begin{figure}[t]
	\centering
	\includegraphics[width=0.99\textwidth]{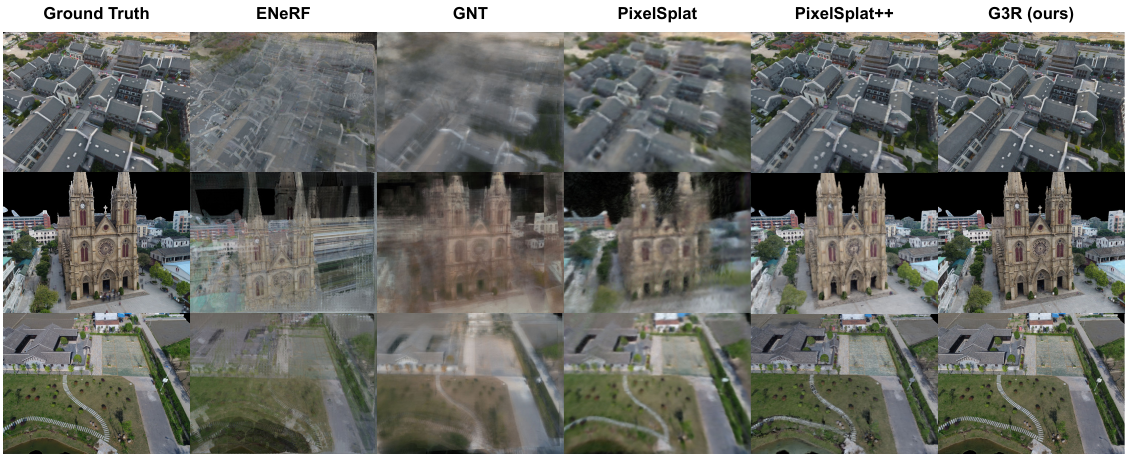}
	\vspace{-0.1in}
	\caption{\textbf{Qualitative comparison to generalizable approaches on BlendedMVS.}}
	\vspace{-0.1in}
	\label{fig:blendedmvs_comp}
\end{figure}

\paragraph{Scene Reconstruction on BlendedMVS:}
We further consider BlendedMVS to evaluate the robustness of different methods to handle many source inputs and large view changes.
As shown in \cref{fig:blendedmvs_comp} and \cref{tab:blendedmvs_comp}, existing generalizable approaches including ENeRF, GNT and PixelSplat cannot handle large view changes and produce bad rendering results with significant visual artifacts due to bad geometry estimation (\eg, blurry appearance, unnatural discontinuity, wrong color palette, etc).
To address this issue, we adapt PixelSplat, named PixelSplat++, to leverage the 3D scaffold to reduce ambiguity and take all available source images for good coverage. Please see Appendix~\ref{sec:baselines} for details.
While achieving signficiant performance boost over existing generalizable methods, PixelSplat++ is still far from per-scene optimization approaches due to the challenge of one-step prediction with limited network capacity.
Our method results in the best photorealism,  minimal reconstruction time and enables real-time rendering speed, which again verifies the effectiveness of our proposed paradigm.
 Moreover, \methodname{} outperforms per-scene optimization methods
especially in perceptual quality. We hypothesize this is because the learned data-driven prior helps handle large view changes better.

 \begin{table}[t]
    \caption{\textbf{Comparison on BlendedMVS.} The methods with best photorealism are marked using gold\raisebox{-0.7ex}{\goldenbullet}, silver\raisebox{-0.7ex}{\graybullet}, and bronze\raisebox{-0.7ex}{\brownbullet} medals. $\dagger$ denotes the method needs to reconstruct the scene again with different source images when rendering each new view.
}
\vspace{-0.13in}

	\resizebox{\textwidth}{!}{
		\setlength{\tabcolsep}{4pt}
		{\renewcommand{\arraystretch}{1.05}
			\begin{tabular}{@{}ll|lll|cc@{}}
				\toprule
				\multicolumn{2}{l|}{\multirow{2}{*}{Models}}   & \multicolumn{3}{c|}{Novel View Synthesis} & \multicolumn{2}{c}{Inference Time} \\
				&                 & PSNR$\uparrow$      & SSIM$\uparrow$     & LPIPS$\downarrow$     & Recon Time           & Render FPS      \\ \midrule
				\multicolumn{1}{l|}{\multirow{4}{*}{Generalizable}}      & ENeRF~\cite{lin2022efficient}           &  15.21  &  0.270  & 0.660 &  0.11s$^\dagger$  &    2.65  \\
				\multicolumn{1}{l|}{} & GNT~\cite{wang2022attention}              &   16.42 &	0.366 &	0.707     &    0.35s$^\dagger$       &  0.00249   \\
				\multicolumn{1}{l|}{} & PixelSplat~\cite{charatan2023pixelsplat}      &   16.24   &   0.344   &   0.781 &  1.14s$^\dagger$  &  176   \\
				\multicolumn{1}{l|}{} & PixelSplat++  &   19.60    &  0.404 &   0.601         &  69s  & 158   \\
				 \midrule
				\multicolumn{1}{l|}{\multirow{2}{*}{Per-scene Opt.}} & Instant-NGP~\cite{mueller2022instant}    &     24.86\raisebox{-0.6ex}{\brownbullet}      &       0.639    &     0.459\raisebox{-0.6ex}{\brownbullet}      &   26min 48s  & 1.65             \\
				\multicolumn{1}{l|}{} & 3DGS~\cite{3dgs}     & 25.12\raisebox{-0.6ex}{\graybullet} &  0.668\raisebox{-0.6ex}{\brownbullet} & 0.462       &   39.5min   & 97.0  \\ \midrule
				\multicolumn{1}{l|}{\multirow{2}{*}{Ours}}    &  \methodname{} (turbo)  & 24.56 & 0.674\raisebox{-0.6ex}{\graybullet} & 0.421\raisebox{-0.6ex}{\graybullet} &  98s&  97.0  \\
				\multicolumn{1}{l|}{} & \methodname{}  &  25.22\raisebox{-0.6ex}{\goldenbullet} &  0.707\raisebox{-0.6ex}{\goldenbullet}   &  0.390\raisebox{-0.6ex}{\goldenbullet}  &   210s  &   97.0 \\ \bottomrule
			\end{tabular}
		}
	}
	\vspace{-0.05in}
	\label{tab:blendedmvs_comp}
\end{table}

\begin{figure}[t]
	\centering
	\includegraphics[width=0.99\textwidth]{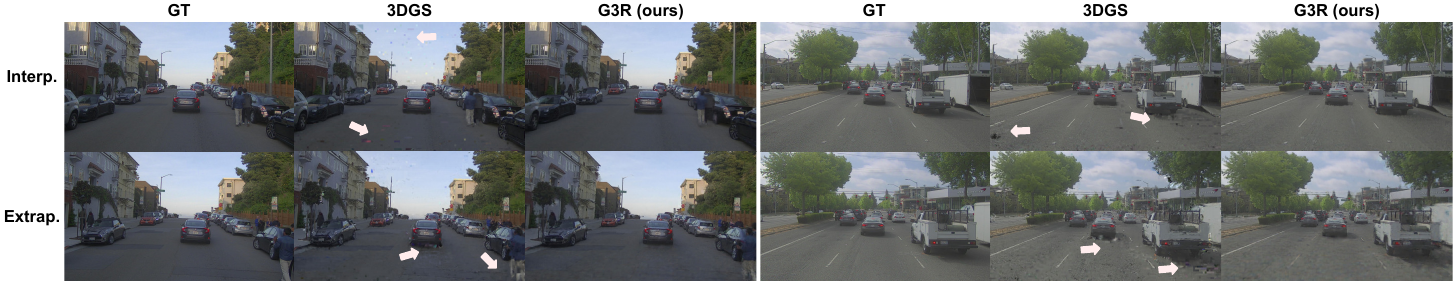}
	\vspace{-0.1in}
	\caption{\textbf{Robustness of G3R vs. 3DGS.}
	3DGS is sharper on interpolation views  (\textit{Interp.}), but has artifacts on extraopolation views  (\textit{Extrap.}).
	}
	\vspace{-0.1in}
	\label{fig:3dgs_comp}
\end{figure}

\paragraph{Robust 3D Gaussian Prediction:}
We compare with the rendering performance of 3DGS at novel views in \cref{fig:3dgs_comp}.
We observe that while 3DGS has sufficient capacity to memorize the source frames, it suffers
a significant performance drop when rendering at novel views due to poor underlying geometry~\cite{guedon2023sugar,cheng2024gaussianpro}.
In contrast, \methodname{} predicts 3D gaussians in a more robust way
{because} \methodname{} is trained with novel view supervision across many scenes (Eq. \ref{equ:loss})
and this supervision helps regularize the 3D neural Gaussians to generalize rather than merely memorize the source views.
We also consider a more challenging \textit{extrapolation} setting where we select 20 consecutive frames as source views and simulate the future 3 frames (\eg , 3 - 6 meters of shift) to evaluate the robustness when rendering at extrapolated views.
As shown in  \cref{fig:3dgs_comp}, \methodname{} results in more realistic rendering performance.
In contrast, 3DGS has severe visual artifacts highlighted by pink arrows (\eg , black holes or wrong colors in road, sky and actor regions).
Please refer to supp. for more analysis.

\paragraph{Ablation study:}

In \cref{tab:ablation}, we ablate the key components proposed in \methodname{} on PandaSet, including replacing the 3D neural Gaussians with the standard 3D Gaussian representation, conducting one-step prediction in both training and inference, training the network only with source view supervision, and switching decaying schedule $\gamma(t)$ to a constant update scale ($0.3$) at each step.
As shown in \cref{tab:ablation}, our proposed neural Gaussian representation is more expressive,
 thus easing the network prediction.
The iterative refinement is critical in the proposed paradigm and single-step prediction fails to generate high-quality reconstruction results.
We notice that single-step G3R is worse than PixelSplat as we enforce smaller updates per step for stable convergence.
Moreover, we show training the network with novel views on many scenes is necessary to enhance the robustness of 3D representation for realistic novel view rendering.
Finally, a proper
update schedule
further improves performance.

\begin{table}[t]
	\begin{minipage}{\textwidth}
		\begin{minipage}[c]{0.55\textwidth}
			\caption{\textbf{Ablation study on PandaSet}.
		}
		\label{tab:ablation}
		\vspace{-0.13in}
		\centering
		\setlength{\tabcolsep}{5pt}
		\resizebox{\textwidth}{!}{
			\begin{tabular}{@{}l|ccc@{}}
	\toprule
	\multicolumn{1}{l|}{\multirow{1}{*}{Models}}    & PSNR      & SSIM      & LPIPS       \\ \midrule
	\multicolumn{1}{l|}{\textbf{Ours}}  &      \textbf{25.22} &   \textbf{0.742}  & \textbf{0.371}                     \\
	\multicolumn{1}{l|}{$-$ 3D neural Gaussian representation}   &  24.72 &  0.718  &  0.420                                 \\
	\multicolumn{1}{l|}{$-$ iterative reconstruction}   & 20.03  &  0.510     &            0.623                             \\
	\multicolumn{1}{l|}{$-$ training with novel views}  &  24.59  & 0.715  &    0.419                     \\
	\multicolumn{1}{l|}{$-$ update schedule $\gamma(t)$}   &  25.03         &    0.732       &        0.400                                  \\
	\bottomrule
\end{tabular}
		}
		\vspace{-0.05in}

	\end{minipage}
	\hfill
	\begin{minipage}[c]{0.45\textwidth}
		\vspace{-0.07in}
		\caption{\textbf{Cross-dataset Generalization.} Pandaset-pretrained model outperforms baselines trained on BlendedMVS (see Tab~\ref{tab:blendedmvs_comp}). }
		\label{tab:cross_dataset}
		\vspace{-0.13in}
		\centering
		\setlength{\tabcolsep}{5pt}
		\resizebox{\textwidth}{!}{
			\begin{tabular}{@{}llll@{}}
				\toprule
				& PSNR & SSIM & LPIPS \\ \midrule
				Zero-short transfer  & 24.11  &  0.653  &   0.448  \\
				Finetune on 2 scenes & 24.99 & 0.676  & 0.428  \\ \bottomrule
			\end{tabular}
		}
	\end{minipage}
\end{minipage}
\end{table}

\paragraph{Generalization study:}
We further evaluate the PandaSet-trained G3R model (static background module) on BlendedMVS (self-driving $\rightarrow$ drone).
The results in Tab.~\ref{tab:cross_dataset} show that G3R trained only on PandaSet achieves significantly better performance
in BlendedMVS than generalizable baselines trained on BlendedMVS directly.
We further finetune the G3R model with only 2 BlendedMVS scenes, achieving comparable results as directly training on full BlendedMVS.
We also showcase applying a Pandaset-pretrained \methodname{} model to
Waymo Open Dataset (WOD)~\cite{waymo} scenes in \cref{fig:waymo}, unveiling the potential for scalable real-world sensor simulation.
See Appendix~\ref{sec:supp_gen_study}. for more analysis.

\begin{figure}[htbp!]
	\centering
	\includegraphics[width=0.99\textwidth]{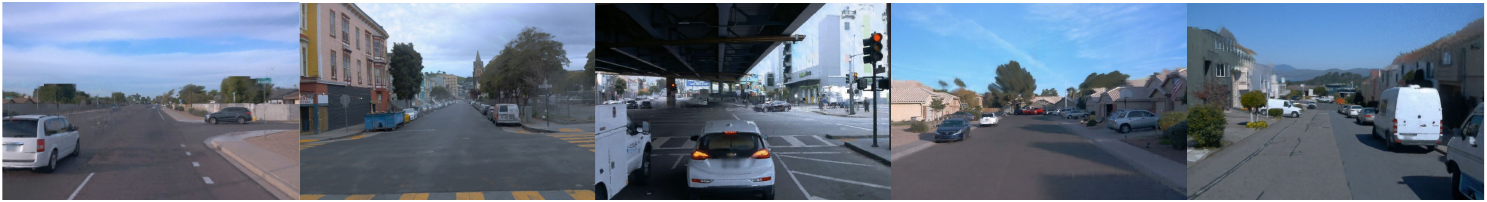}
	\vspace{-0.1in}
	\caption{\textbf{PandaSet-pretrained model generalizes to Waymo Open Dataset.}} 
	\vspace{-0.1in}
	\label{fig:waymo}
\end{figure}

\begin{figure}[t]
	\centering
	\includegraphics[width=0.99\textwidth]{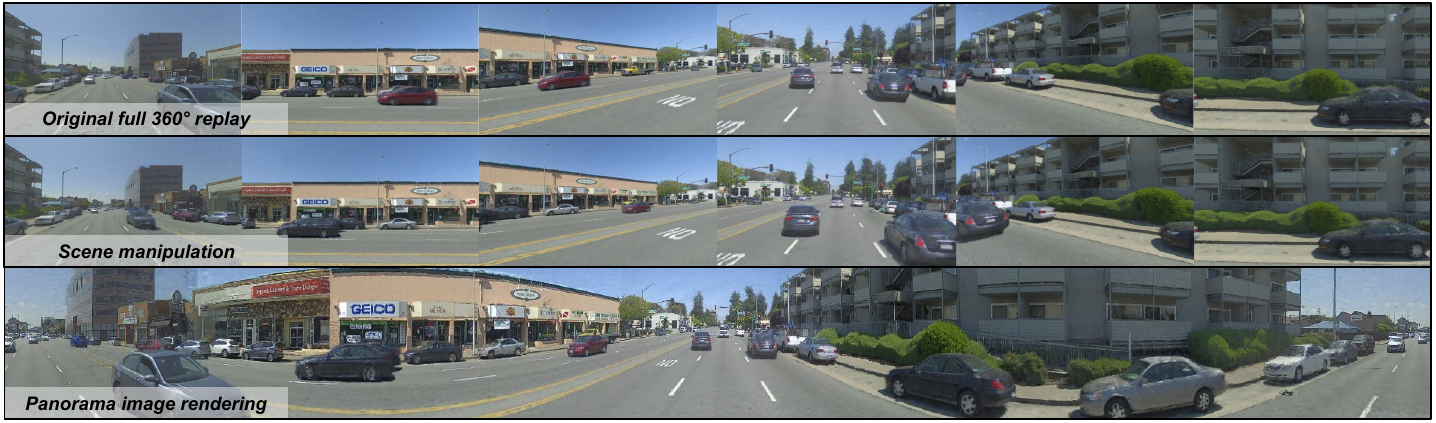}
	\vspace{-0.1in}
	\caption{\textbf{Realistic and controllable multi-camera simulation on PandaSet.}
	\methodname{} reconstructs a manipulable 3D scene representation.
	}
	\vspace{-0.1in}
	\label{fig:camerasim_demo}
\end{figure}

\paragraph{Realistic and controllable camera simulation:}
We now showcase applying \methodname{} for high-fidelity multi-camera simulation in large-scale driving scenarios.
Compared to previous generalizable approaches, our method can reconstruct a standalone representation, which allows us to control, edit and interactively render the scene for various applications. %
In \cref{fig:camerasim_demo}, we show \methodname{}-reconstructed scene can synthesize consistent and high-fidelity multi-camera videos from one single driving pass (top row). Moreover, we can manipulate the scene by freezing the sensors and changing the positions of dynamic actors,
and render corresponding multi-camera (second row) or panorama images (bottom row).

\paragraph{Limitations:}
Our approach has artifacts in large extrapolations, which may require scene completion.
Better surface regularization~\cite{guedon2023sugar,cheng2024gaussianpro} and adversarial training~\cite{roessle2023ganerf,unisim} may
mitigate these issues.
\methodname{}'s performance suffers when initialized with sparse points, but can leverage LiDAR or fast MVS techniques~\cite{wu2024gomvs} to mitigate this.
We also do not model non-rigid deformations \cite{luiten2023dynamic} and emissive lighting. %
See Appendix~\ref{sec:limitations} for details.

\section{Conclusion}
In this paper, we introduce \methodname{}, a novel approach for efficient generalizable large-scale 3D scene reconstruction.
By leveraging gradient feedback signals from differentiable rendering, \methodname{} achieves acceleration of at
least $10 \times$
over state-of-the-art per-scene optimization methods,
with comparable or superior photorealism.
Importantly, our method predicts a standalone 3D representation that exhibits robustness to large view changes and enables real-time rendering, making it well-suited for VR and
simulation. Experiments on urban-driving and drone datasets showcase the efficacy of \methodname{} for in-the-wild 3D scene reconstruction.
Our learning-to-optimize paradigm with gradient signal can apply
to other 3D representations such as triplanes with NeRF rendering, or other inverse problems such as generalizable surface reconstruction~\cite{long22sparseneus, ren2022volrecon, liang2023retr, huang2023nksr}.

\section*{Acknowledgement}
We sincerely thank the anonymous reviewers for their insightful comments and suggestions. We thank the Waabi team for their valuable assistance and support.

\bibliographystyle{splncs04}
\bibliography{ref_clean}

\begin{thebibliography}{10}
\providecommand{\url}[1]{\texttt{#1}}
\providecommand{\urlprefix}{URL }
\providecommand{\doi}[1]{https://doi.org/#1}

\bibitem{altizure}
Altizure: Mapping the world in 3d. https://www. altizure.com

\bibitem{adler2017solving}
Adler, J., Öktem, O.: Solving ill-posed inverse problems using iterative deep
  neural networks. arXiv  (2017)

\bibitem{aliev2020neural}
Aliev, K.A., Sevastopolsky, A., Kolos, M., Ulyanov, D., Lempitsky, V.: Neural
  point-based graphics. In: ECCV (2020)

\bibitem{andrychowicz2016learning}
Andrychowicz, M., Denil, M., Colmenarejo, S.G., Hoffman, M.W., Pfau, D.,
  Schaul, T., de~Freitas, N.: Learning to learn by gradient descent by gradient
  descent. NeurIPS  (2016)

\bibitem{carreira2015human}
Carreira, J., Agrawal, P., Fragkiadaki, K., Malik, J.: Human pose estimation
  with iterative error feedback. CVPR  (2015)

\bibitem{charatan2023pixelsplat}
Charatan, D., Li, S., Tagliasacchi, A., Sitzmann, V.: pixelsplat: 3d gaussian
  splats from image pairs for scalable generalizable 3d reconstruction. arXiv
  (2023)

\bibitem{chen2021mvsnerf}
Chen, A., Xu, Z., Zhao, F., Zhang, X., Xiang, F., Yu, J., Su, H.: Mvsnerf: Fast
  generalizable radiance field reconstruction from multi-view stereo. In: ICCV
  (2021)

\bibitem{cheng2024gaussianpro}
Cheng, K., Long, X., Yang, K., Yao, Y., Yin, W., Ma, Y., Wang, W., Chen, X.:
  Gaussianpro: 3d gaussian splatting with progressive propagation. arXiv
  (2024)

\bibitem{chibane2021stereo}
Chibane, J., Bansal, A., Lazova, V., Pons-Moll, G.: Stereo radiance fields
  (srf): Learning view synthesis for sparse views of novel scenes. CVPR  (2021)

\bibitem{cciccek20163d}
{\c{C}}i{\c{c}}ek, {\"O}., Abdulkadir, A., Lienkamp, S.S., Brox, T.,
  Ronneberger, O.: 3d u-net: learning dense volumetric segmentation from sparse
  annotation. In: Medical Image Computing and Computer-Assisted
  Intervention--MICCAI 2016: 19th International Conference, Athens, Greece,
  October 17-21, 2016, Proceedings, Part II 19 (2016)

\bibitem{cong2023enhancing}
Cong, W., Liang, H., Wang, P., Fan, Z., Chen, T., Varma, M., Wang, Y., Wang,
  Z.: Enhancing nerf akin to enhancing llms: Generalizable nerf transformer
  with mixture-of-view-experts. In: ICCV (2023)

\bibitem{flynn2019deepview}
Flynn, J., Broxton, M., Debevec, P., DuVall, M., Fyffe, G., Overbeck, R.,
  Snavely, N., Tucker, R.: Deepview: View synthesis with learned gradient
  descent. In: CVPR (2019)

\bibitem{guedon2023sugar}
Guédon, A., Lepetit, V.: Sugar: Surface-aligned gaussian splatting for
  efficient 3d mesh reconstruction and high-quality mesh rendering. arXiv
  (2023)

\bibitem{hong2024lrm}
Hong, Y., Zhang, K., Gu, J., Bi, S., Zhou, Y., Liu, D., Liu, F., Sunkavalli,
  K., Bui, T., Tan, H.: {LRM}: Large reconstruction model for single image to
  3d. In: The Twelfth International Conference on Learning Representations
  (2024), \url{https://openreview.net/forum?id=sllU8vvsFF}

\bibitem{hu2019taichi}
Hu, Y., Li, T.M., Anderson, L., Ragan-Kelley, J., Durand, F.: Taichi: a
  language for high-performance computation on spatially sparse data
  structures. ACM Transactions on Graphics (TOG)  \textbf{38}(6), ~201 (2019)

\bibitem{huang2023nksr}
Huang, J., Gojcic, Z., Atzmon, M., Litany, O., Fidler, S., Williams, F.: Neural
  kernel surface reconstruction. CVPR  (2023)

\bibitem{huang2023neural}
Huang, S., Gojcic, Z., Wang, Z., Williams, F., Kasten, Y., Fidler, S.,
  Schindler, K., Litany, O.: Neural lidar fields for novel view synthesis.
  arXiv  (2023)

\bibitem{johari2022geonerf}
Johari, M.M., Lepoittevin, Y., Fleuret, F.: Geonerf: Generalizing nerf with
  geometry priors. In: CVPR (2022)

\bibitem{3dgs}
Kerbl, B., Kopanas, G., Leimk{\"u}hler, T., Drettakis, G.: 3d gaussian
  splatting for real-time radiance field rendering. TOG  (2023)

\bibitem{kingma2014adam}
Kingma, D.P., Ba, J.: Adam: A method for stochastic optimization. ICLR  (2015)

\bibitem{kopanas2021pointbased}
Kopanas, G., Philip, J., Leimkühler, T., Drettakis, G.: Point-based neural
  rendering with per-view optimization. Computer graphics forum (Print)  (2021)

\bibitem{kulhanek2022viewformer}
Kulh'anek, J., Derner, E., Sattler, T., Babuvska, R.: Viewformer: Nerf-free
  neural rendering from few images using transformers. ECCV  (2022)

\bibitem{li2024instantd}
Li, J., Tan, H., Zhang, K., Xu, Z., Luan, F., Xu, Y., Hong, Y., Sunkavalli, K.,
  Shakhnarovich, G., Bi, S.: Instant3d: Fast text-to-3d with sparse-view
  generation and large reconstruction model. In: The Twelfth International
  Conference on Learning Representations (2024),
  \url{https://openreview.net/forum?id=2lDQLiH1W4}

\bibitem{li2016learning}
Li, K., Malik, J.: Learning to optimize. ICLR  (2016)

\bibitem{li2018deepim}
Li, Y., Wang, G., Ji, X., Xiang, Y., Fox, D.: Deepim: Deep iterative matching
  for 6d pose estimation. IJCV  (2018)

\bibitem{liang2023retr}
Liang, Y., He, H., Chen, Y.: Retr: Modeling rendering via transformer for
  generalizable neural surface reconstruction. NeurIPS  (2023)

\bibitem{lin2022efficient}
Lin, H., Peng, S., Xu, Z., Yan, Y., Shuai, Q., Bao, H., Zhou, X.: Efficient
  neural radiance fields for interactive free-viewpoint video. In: SIGGRAPH
  Asia 2022 Conference Papers (2022)

\bibitem{lin2024vastgaussian}
Lin, J., Li, Z., Tang, X., Liu, J., Liu, S., Liu, J., Lu, Y., Wu, X., Xu, S.,
  Yan, Y., et~al.: Vastgaussian: Vast 3d gaussians for large scene
  reconstruction. arXiv  (2024)

\bibitem{lin2023urbanir}
Lin, Z.H., Liu, B., Chen, Y.T., Forsyth, D., Huang, J.B., Bhattad, A., Wang,
  S.: Urbanir: Large-scale urban scene inverse rendering from a single video.
  arXiv  (2023)

\bibitem{liu2023real}
Liu, J.Y., Chen, Y., Yang, Z., Wang, J., Manivasagam, S., Urtasun, R.:
  Real-time neural rasterization for large scenes. In: ICCV (2023)

\bibitem{liu2023zero}
Liu, R., Wu, R., Van~Hoorick, B., Tokmakov, P., Zakharov, S., Vondrick, C.:
  Zero-1-to-3: Zero-shot one image to 3d object. In: Proceedings of the
  IEEE/CVF international conference on computer vision. pp. 9298--9309 (2023)

\bibitem{liu2022neural}
Liu, Y., Peng, S., Liu, L., Wang, Q., Wang, P., Theobalt, C., Zhou, X., Wang,
  W.: Neural rays for occlusion-aware image-based rendering. In: CVPR (2022)

\bibitem{ljungbergh2024neuroncap}
Ljungbergh, W., Tonderski, A., Johnander, J., Caesar, H., {\AA}str{\"o}m, K.,
  Felsberg, M., Petersson, C.: Neuroncap: Photorealistic closed-loop safety
  testing for autonomous driving. arXiv preprint arXiv:2404.07762  (2024)

\bibitem{long22sparseneus}
Long, X., Lin, C., Wang, P., Komura, T., Wang, W.: Sparseneus: Fast
  generalizable neural surface reconstruction from sparse views. In: ECCV
  (2022)

\bibitem{luiten2023dynamic}
Luiten, J., Kopanas, G., Leibe, B., Ramanan, D.: Dynamic 3d gaussians: Tracking
  by persistent dynamic view synthesis. arXiv  (2023)

\bibitem{ma2021deep}
Ma, W.C., Wang, S., Gu, J., Manivasagam, S., Torralba, A., Urtasun, R.: Deep
  feedback inverse problem solver. ECCV  (2021)

\bibitem{manhardt2018deep}
Manhardt, F., Kehl, W., Navab, N., Tombari, F.: Deep model-based 6d pose
  refinement in rgb. ECCV  (2018)

\bibitem{manivasagam2023towards}
Manivasagam, S., B{\^a}rsan, I.A., Wang, J., Yang, Z., Urtasun, R.: Towards
  zero domain gap: A comprehensive study of realistic lidar simulation for
  autonomy testing. In: {ICCV} (2023)

\bibitem{mildenhall2020nerf}
Mildenhall, B., Srinivasan, P.P., Tancik, M., Barron, J.T., Ramamoorthi, R.,
  Ng, R.: Nerf: Representing scenes as neural radiance fields for view
  synthesis. ECCV  (2020)

\bibitem{mueller2022instant}
M\"uller, T., Evans, A., Schied, C., Keller, A.: Instant neural graphics
  primitives with a multiresolution hash encoding  (2022)

\bibitem{mller2022autorf}
Müller, N., Simonelli, A., Porzi, L., Bulò, S.R., Nießner, M., Kontschieder,
  P.: Autorf: Learning 3d object radiance fields from single view observations.
  CVPR  (2022)

\bibitem{niemeyer2020differentiable}
Niemeyer, M., Mescheder, L., Oechsle, M., Geiger, A.: Differentiable volumetric
  rendering: Learning implicit 3d representations without 3d supervision. In:
  CVPR (2020)

\bibitem{oquab2023dinov2}
Oquab, M., Darcet, T., Moutakanni, T., Vo, H., Szafraniec, M., Khalidov, V.,
  Fernandez, P., Haziza, D., Massa, F., El-Nouby, A., et~al.: Dinov2: Learning
  robust visual features without supervision. arXiv  (2023)

\bibitem{ost2021neural}
Ost, J., Mannan, F., Thuerey, N., Knodt, J., Heide, F.: Neural scene graphs for
  dynamic scenes. CVPR  (2021)

\bibitem{pun2023neural}
Pun, A., Sun, G., Wang, J., Chen, Y., Yang, Z., Manivasagam, S., Ma, W.C.,
  Urtasun, R.: Neural lighting simulation for urban scenes. In: NeurIPS (2023)

\bibitem{reizenstein2021common}
Reizenstein, J., Shapovalov, R., Henzler, P., Sbordone, L., Labatut, P.,
  Novotný, D.: Common objects in 3d: Large-scale learning and evaluation of
  real-life 3d category reconstruction. ICCV  (2021)

\bibitem{ren2022volrecon}
Ren, Y., Wang, F., Zhang, T., Pollefeys, M., Susstrunk, S.E.: Volrecon: Volume
  rendering of signed ray distance functions for generalizable multi-view
  reconstruction. CVPR  (2022)

\bibitem{riegler2020fvs}
Riegler, G., Koltun, V.: Free view synthesis. ECCV  (2020)

\bibitem{riegler2021svs}
Riegler, G., Koltun, V.: Stable view synthesis. CVPR  (2021)

\bibitem{roessle2023ganerf}
Roessle, B., M{\"u}ller, N., Porzi, L., Bul{\`o}, S.R., Kontschieder, P.,
  Nie{\ss}ner, M.: Ganerf: Leveraging discriminators to optimize neural
  radiance fields. ACM Trans. Graph.  (2023)

\bibitem{rombach2021geometryfree}
Rombach, R., Esser, P., Ommer, B.: Geometry-free view synthesis: Transformers
  and no 3d priors. ICCV  (2021)

\bibitem{sajjadi2022rust}
Sajjadi, M.S.M., Mahendran, A., Kipf, T., Pot, E., Duckworth, D., Lucic, M.,
  Greff, K.: Rust: Latent neural scene representations from unposed imagery.
  CVPR  (2022)

\bibitem{srt22}
Sajjadi, M.S.M., Meyer, H., Pot, E., Bergmann, U., Greff, K., Radwan, N., Vora,
  S., Lucic, M., Duckworth, D., Dosovitskiy, A., Uszkoreit, J., Funkhouser, T.,
  Tagliasacchi, A.: {Scene Representation Transformer: Geometry-Free Novel View
  Synthesis Through Set-Latent Scene Representations}. CVPR  (2022)

\bibitem{sarva2023adv3d}
Sarva, J., Wang, J., Tu, J., Xiong, Y., Manivasagam, S., Urtasun, R.: Adv3d:
  Generating safety-critical 3d objects through closed-loop simulation. In: 7th
  Annual Conference on Robot Learning (2023),
  \url{https://openreview.net/forum?id=nyY6UgXYyfF}

\bibitem{seitzer2023dyst}
Seitzer, M., van Steenkiste, S., Kipf, T., Greff, K., Sajjadi, M.S.M.: Dyst:
  Towards dynamic neural scene representations on real-world videos. arXiv
  (2023)

\bibitem{sitzmann2021light}
Sitzmann, V., Rezchikov, S., Freeman, W., Tenenbaum, J., Durand, F.: Light
  field networks: Neural scene representations with single-evaluation
  rendering. NeurIPS  (2021)

\bibitem{SitzmannZW19SRN}
Sitzmann, V., Zollh{\"{o}}fer, M., Wetzstein, G.: Scene representation
  networks: Continuous 3d-structure-aware neural scene representations. In:
  NeurIPS (2019)

\bibitem{snavely2006phototourism}
Snavely, N., Seitz, S.M., Szeliski, R.: Photo tourism: Exploring photo
  collections in 3d. SIGGRAPH  (2006)

\bibitem{song2020denoising}
Song, J., Meng, C., Ermon, S.: Denoising diffusion implicit models. ICLR
  (2020)

\bibitem{srinivasan2019pushing}
Srinivasan, P.P., Tucker, R., Barron, J.T., Ramamoorthi, R., Ng, R., Snavely,
  N.: Pushing the boundaries of view extrapolation with multiplane images.
  arXiv  (2019)

\bibitem{suhail2021light}
Suhail, M., Esteves, C., Sigal, L., Makadia, A.: Light field neural rendering.
  CVPR  (2021)

\bibitem{suhail2022generalizable}
Suhail, M., Esteves, C., Sigal, L., Makadia, A.: Generalizable patch-based
  neural rendering. In: ECCV (2022)

\bibitem{waymo}
Sun, P., Kretzschmar, H., Dotiwalla, X., Chouard, A., Patnaik, V., Tsui, P.,
  Guo, J., Zhou, Y., Chai, Y., Caine, B., Vasudevan, V., Han, W., Ngiam, J.,
  Zhao, H., Timofeev, A., Ettinger, S., Krivokon, M., Gao, A., Joshi, A.,
  Zhang, Y., Shlens, J., Chen, Z., Anguelov, D.: Scalability in perception for
  autonomous driving: Waymo open dataset. In: CVPR (2020)

\bibitem{tancik2022block}
Tancik, M., Casser, V., Yan, X., Pradhan, S., Mildenhall, B., Srinivasan, P.P.,
  Barron, J.T., Kretzschmar, H.: Block-nerf: Scalable large scene neural view
  synthesis. In: CVPR (2022)

\bibitem{tangandyang2023torchsparse}
Tang, H., Yang, S., Liu, Z., Hong, K., Yu, Z., Li, X., Dai, G., Wang, Y., Han,
  S.: Torchsparse++: Efficient training and inference framework for sparse
  convolution on gpus. In: IEEE/ACM International Symposium on
  Microarchitecture (MICRO) (2023)

\bibitem{tonderski2023neurad}
Tonderski, A., Lindstr{\"o}m, C., Hess, G., Ljungbergh, W., Svensson, L.,
  Petersson, C.: Neurad: Neural rendering for autonomous driving. arXiv  (2023)

\bibitem{trevithick2021grf}
Trevithick, A., Yang, B.: Grf: Learning a general radiance field for 3d
  representation and rendering. In: ICCV (2021)

\bibitem{turki2022mega}
Turki, H., Ramanan, D., Satyanarayanan, M.: Mega-nerf: Scalable construction of
  large-scale nerfs for virtual fly-throughs. In: CVPR (2022)

\bibitem{wang2022cadsim}
Wang, J., Manivasagam, S., Chen, Y., Yang, Z., B{\^a}rsan, I.A., Yang, A.J.,
  Ma, W.C., Urtasun, R.: {CADS}im: Robust and scalable in-the-wild 3d
  reconstruction for controllable sensor simulation. In: 6th Annual Conference
  on Robot Learning (2022)

\bibitem{wang2021advsim}
Wang, J., Pun, A., Tu, J., Manivasagam, S., Sadat, A., Casas, S., Ren, M.,
  Urtasun, R.: Advsim: Generating safety-critical scenarios for self-driving
  vehicles. In: CVPR (2021)

\bibitem{wang2022attention}
Wang, P., Chen, X., Chen, T., Venugopalan, S., Wang, Z., et~al.: Is attention
  all nerf needs? arXiv  (2022)

\bibitem{wang2021ibrnet}
Wang, Q., Wang, Z., Genova, K., Srinivasan, P.P., Zhou, H., Barron, J.T.,
  Martin-Brualla, R., Snavely, N., Funkhouser, T.: Ibrnet: Learning multi-view
  image-based rendering. In: CVPR (2021)

\bibitem{wang2004image}
Wang, Z., Bovik, A.C., Sheikh, H.R., Simoncelli, E.P.: Image quality
  assessment: from error visibility to structural similarity. TIP  (2004)

\bibitem{wang2023neural}
Wang, Z., Shen, T., Gao, J., Huang, S., Munkberg, J., Hasselgren, J., Gojcic,
  Z., Chen, W., Fidler, S.: Neural fields meet explicit geometric
  representations for inverse rendering of urban scenes. In: CVPR (2023)

\bibitem{wei2024meshlrm}
Wei, X., Zhang, K., Bi, S., Tan, H., Luan, F., Deschaintre, V., Sunkavalli, K.,
  Su, H., Xu, Z.: Meshlrm: Large reconstruction model for high-quality mesh.
  arXiv preprint arXiv:2404.12385  (2024)

\bibitem{wichrowska2017learned}
Wichrowska, O., Maheswaranathan, N., Hoffman, M.W., Colmenarejo, S.G., Denil,
  M., de~Freitas, N., Sohl-Dickstein, J.N.: Learned optimizers that scale and
  generalize. ICML  (2017)

\bibitem{wiles2019synsin}
Wiles, O., Gkioxari, G., Szeliski, R., Johnson, J.: Synsin: End-to-end view
  synthesis from a single image. arXiv  (2019)

\bibitem{wu2024gomvs}
Wu, J., Li, R., Xu, H., Zhao, W., Zhu, Y., Sun, J., Zhang, Y.: Gomvs:
  Geometrically consistent cost aggregation for multi-view stereo. In: CVPR
  (2024)

\bibitem{wu2023mars}
Wu, Z., Liu, T., Luo, L., Zhong, Z., Chen, J., Xiao, H., Hou, C., Lou, H.,
  Chen, Y., Yang, R., et~al.: Mars: An instance-aware, modular and realistic
  simulator for autonomous driving. arXiv  (2023)

\bibitem{xiao2021pandaset}
Xiao, P., Shao, Z., Hao, S., Zhang, Z., Chai, X., Jiao, J., Li, Z., Wu, J.,
  Sun, K., Jiang, K., et~al.: Pandaset: Advanced sensor suite dataset for
  autonomous driving. In: ITSC (2021)

\bibitem{xiong2023learning}
Xiong, Y., Ma, W.C., Wang, J., Urtasun, R.: Learning compact representations
  for lidar completion and generation. In: CVPR (2023)

\bibitem{yan2024street}
Yan, Y., Lin, H., Zhou, C., Wang, W., Sun, H., Zhan, K., Lang, X., Zhou, X.,
  Peng, S.: Street gaussians for modeling dynamic urban scenes. arXiv  (2024)

\bibitem{yang2023contranerf}
Yang, H., Hong, L., Li, A., Hu, T., Li, Z., Lee, G.H., Wang, L.: Contranerf:
  Generalizable neural radiance fields for synthetic-to-real novel view
  synthesis via contrastive learning. In: CVPR (2023)

\bibitem{yang2023emernerf}
Yang, J., Ivanovic, B., Litany, O., Weng, X., Kim, S.W., Li, B., Che, T., Xu,
  D., Fidler, S., Pavone, M., et~al.: Emernerf: Emergent spatial-temporal scene
  decomposition via self-supervision. arXiv  (2023)

\bibitem{unisim}
Yang, Z., Chen, Y., Wang, J., Manivasagam, S., Ma, W.C., Yang, A.J., Urtasun,
  R.: Unisim: A neural closed-loop sensor simulator. In: CVPR (2023)

\bibitem{yang2023reconstructing}
Yang, Z., Manivasagam, S., Chen, Y., Wang, J., Hu, R., Urtasun, R.:
  Reconstructing objects in-the-wild for realistic sensor simulation. In: ICRA
  (2023)

\bibitem{yang2021recovering}
Yang, Z., Manivasagam, S., Liang, M., Yang, B., Ma, W.C., Urtasun, R.:
  Recovering and simulating pedestrians in the wild. In: Conference on Robot
  Learning. pp. 419--431. PMLR (2021)

\bibitem{yang2021s3}
Yang, Z., Wang, S., Manivasagam, S., Huang, Z., Ma, W.C., Yan, X., Yumer, E.,
  Urtasun, R.: S3: Neural shape, skeleton, and skinning fields for 3d human
  modeling. In: CVPR. pp. 13284--13293 (2021)

\bibitem{yao2019blendedmvs}
Yao, Y., Luo, Z., Li, S., Zhang, J., Ren, Y., Zhou, L., Fang, T., Quan, L.:
  Blendedmvs: A large-scale dataset for generalized multi-view stereo networks.
  CVPR  (2020)

\bibitem{yu2021pixelnerf}
Yu, A., Ye, V., Tancik, M., Kanazawa, A.: pixelnerf: Neural radiance fields
  from one or few images. In: CVPR (2021)

\bibitem{zhang2024gs}
Zhang, K., Bi, S., Tan, H., Xiangli, Y., Zhao, N., Sunkavalli, K., Xu, Z.:
  Gs-lrm: Large reconstruction model for 3d gaussian splatting. arXiv preprint
  arXiv:2404.19702  (2024)

\bibitem{zhang2018unreasonable}
Zhang, R., Isola, P., Efros, A.A., Shechtman, E., Wang, O.: The unreasonable
  effectiveness of deep features as a perceptual metric. CVPR  (2018)

\bibitem{zhenxing2022switch}
Zhenxing, M., Xu, D.: Switch-nerf: Learning scene decomposition with mixture of
  experts for large-scale neural radiance fields. In: ICLR (2022)

\bibitem{zhou2018stereo}
Zhou, T., Tucker, R., Flynn, J., Fyffe, G., Snavely, N.: Stereo magnification:
  Learning view synthesis using multiplane images. SIGGRAPH  (2018)

\bibitem{zhou2023drivinggaussian}
Zhou, X., Lin, Z., Shan, X., Wang, Y., Sun, D., Yang, M.H.: Drivinggaussian:
  Composite gaussian splatting for surrounding dynamic autonomous driving
  scenes. arXiv  (2023)

\end{thebibliography}

\newpage
\appendix
\renewcommand{\thetable}{A\arabic{table}}
\renewcommand{\thefigure}{A\arabic{figure}}
\renewcommand{\thealgorithm}{A\arabic{algorithm}}

\section*{Appendix}
In this appendix, we provide additional information on \methodname{},
experimental setup, additional quantitative and qualitative results, limitations, and broader implications.
We first provide additional information and motivation on
\methodname{} (Sec~\ref{sec:details}).
In Sec.~\ref{sec:baselines}, we provide details on baseline implementations and how we adapt them to urban-driving and drone datasets.
Next, we provide the experimental setup for evaluation on urban-driving and drone datasets in Sec.~\ref{sec:exp_setup}.
We then show more qualitative comparison with baselines (Sec.~\ref{sec:qualitative_supp}), multi-camera simulation results (Sec.~\ref{sec:supp_controllable_sim}) and a generalization study across datasets (Sec.~\ref{sec:supp_gen_study}).
Finally, we discuss the limitations (Sec.~\ref{sec:limitations}) and broader impact (Sec.~\ref{sec:broader_impact}).

\section{G3R Implementation Details}
\label{sec:details}

We first discuss three major paradigms for scene reconstruction as shown in main-paper-Fig. 2 and then present implementation details for \methodname{}.

\subsection{Comparison of Three Paradigms for Scene Reconstruction}
For better understanding, we provide detailed algorithms for three paradigms for scene reconstruction discussed in the main paper.
Each algorithm box depicts the paradigm's approach to reconstruct a new scene at inference time.

\begin{algorithm}[htbp!]
	\caption{Generalizable Novel View Synthesis}
	\label{alg:gen_nvs}
	\begin{algorithmic}
		\State \textbf{Inputs:} Source Images  $\rmI^{\mathrm{src}}$,  target view $\mathrm{\Pi}^{\mathrm{tgt}}$, reconstruction encoder
		$G_\theta$, decoder network $D_\theta: \mathcal{S} \to \mathbf I$
		\State $\rmI_{\mathrm{nn}}^{\mathrm{src}}$ = Select$(\rmI^{\mathrm{src}}, \mathrm{\Pi}^{\mathrm{tgt}})$ \hfill \# select nearest neighboring source views
		\State $\mathcal S_\text{nn} \gets
		G_\theta(\rmI_{\mathrm{nn}}^{\mathrm{src}}, \mathrm{\Pi}^{\mathrm{tgt}})$ \hfill  \# predicted representation depends on view selection
		\State $\hat{\rmI}^{\mathrm{tgt}} = D_\theta (\mathcal{S}_\text{nn}, \mathrm{\Pi}^{\mathrm{tgt}})$  \hfill \# render single target image from target view
		\State \textbf{Return} $\mathcal{S}_\text{nn}$ \hfill \# only renders views close to $\mathrm{\Pi}^{\mathrm{tgt}}$, need to re-run if it changes
	\end{algorithmic}
\end{algorithm}

Algorithm~\ref{alg:gen_nvs} and~\ref{alg:per-scene-opt} show the generalizable novel view synthesis (Fig. 2a) and per-scene optimization paradigms (Fig. 2b) separately. Specifically, existing generalizable approaches select a few reference images (usually $\leq 5$) for feed-forward prediction of intermediate representation and then decode/render the feature representation to produce the rendered images. These approaches learn data-driven priors across multiple scenes and enable fast reconstruction.
They need to reconstruct the scene again with different source images when rendering a new view. Existing generalizable approaches work only for small objects/scenes and small view changes due to limited network capacity and handle a small number of source images due to memory constraints.

Recently, neural rendering approaches such as NeRF and 3D Gaussian Splatting have achieved realistic reconstructions for large scenes. These methods take all source images and reconstruct a 3D representation via energy minimization and differentiable rendering to the source views. However, they require a costly per-scene optimization process which usually takes several hours ($T > 1000$) and often exhibit artifacts when the view changes are large due to overfitting.

\begin{algorithm}[t]
	\caption{Per-Scene Reconstruction by Gradient-Descent}
	\label{alg:per-scene-opt}
	\begin{algorithmic}
		\State \textbf{Input:} Initial scene representation $\mathcal S^{(0)}$,  source images  $\mathbf I^{\mathrm{src}}$,  renderer $f_{\mathrm{render}}: \mathcal S\to \mathbf I	$, optimiaztion iterations $T$ (usually > 1000)
		\For{$t = 0,1,2, \dots, T - 1$}
		\State $\hat{\rmI}^{\mathrm{src}} = f_\mathrm{render}(\mathcal{S}^{(t)})$
		\State $\nabla_{\mathcal{S}^{(t)}} \gets \nabla \lVert \rmI^{\mathrm{src}}-\hat{\rmI}^{\mathrm{src}} \rVert_2$
		\State $\mathcal S^{(t+1)} = \mathcal S^{(t)} - \nabla_{\mathcal S^{(t)}}$
		\EndFor
		\State \textbf{Return} $\mathcal{S}^{(T)}$
	\end{algorithmic}
\end{algorithm}

\begin{algorithm}[t]
	\caption{Gradient-Guided Generalizable Reconstruction (G3R)}
	\label{alg:g3r}
	\begin{algorithmic}
		\State \textbf{Input:} Initial scene representation $\mathcal S^{(0)}$, source Images  $\mathbf I^{\mathrm{src}}$, renderer $f_{\mathrm{render}}: \mathcal S\to \rmI$,  \textcolor{red}{reconstruction network $G_\theta$}, \textcolor{red}{update iterations $T = 24$}
		\For{$t = 0,1,2, \dots, \textcolor{red}{24}$}
		\State $\hat{\rmI}^{\mathrm{src}} = f_{\mathrm{render}}(\mathcal S^{(t)})$
		\State $\nabla_{\mathcal{S}^{(t)}} \gets \nabla \lVert \rmI^{\mathrm{src}}-\hat{\rmI}^{\mathrm{src}} \rVert_2$  \hfill \# lift 2D to 3D as gradients
		\State $\mathcal S^{(t+1)} = \mathcal S^{(t)} + \gamma (t) \cdot \textcolor{red}{G_\theta(\mathcal S^{(t)}, \nabla_{\mathcal S^{(t)}}; t)} $  \hfill \# iteratively refine the 3D representation
		\EndFor
		\State \textbf{Return}  $\mathcal S^{(T)}$
	\end{algorithmic}
\end{algorithm}

To enable fast large scene reconstruction while achieving high-fidity rendering performance, we instead propose to learn a network that iteratively refines a 3D scene representation with 3D gradient guidance (Algorithm~\ref{alg:g3r}). We highlight the major differences of G3R paradigm compared to the other two paradigms \textcolor{red}{in red}.
Our key idea is to learn a single reconstruction network that iteratively updates the 3D scene representation, combining the benefits of data-driven priors from fast prediction methods with the iterative gradient feedback signal from per-scene optimization methods.
\methodname{} can be viewed as a ``learned optimizer''  that leverages spatial correlation and data-driven priors for fast scene reconstruction.

\subsection{G3R Training Algorithm}
We further show the presudocode algorithm for G3R reconstruction network training in Algoirthm~\ref{alg:g3r_training} (See Eqn.~1, 4 and 6). G3R-Net takes current 3D neural Gaussians $\mathcal{S}^{(t)}$ and 3D gradient $\nabla_{\mathcal{S}^{(t)}}$ and output the refinement $\Delta \mathcal{S}^{(t)}$. We update the parameters of the reconstruction network and transformation MLP at every update step $t$.

\begin{algorithm}[htbp!]
	\caption{G3R-Net Training}
	\label{alg:g3r_training}
	\begin{algorithmic}
		\State \textbf{Input}: Data $\mathcal{D}$: collection of (scene $\mathcal{S}^{(0)}$, images $\rm\bf I$, poses $\mathrm{\Pi}$) pairs, $f_{\mathrm{rast}}$: differential tile renderer, $G_{\theta}$: generalizable reconstruction network,  $f_{\mathrm{mlp}}$: transformation MLP, $\gamma(t)$ update scheduler
		\While{$G_\theta$ not converged}
		\State $\mathcal S^{(0)}, \rmI, \mathrm{\Pi} = \text{Sample}(\mathcal{D})$
		\State $(\rmI^{\mathrm{src}}, \mathrm{\Pi}^{\mathrm{src}}), (\rmI^{\mathrm{tgt}}, \mathrm{\Pi}^{\mathrm{tgt}}) = \text{Split}(\rmI, \mathrm{\Pi})$
		\For{$t = 0, 1, 2, \dots,T - 1$}
		\State $\nabla_{\mathcal{S}^{(t)}} = \nabla \lVert\rmI^{\mathrm{src}} - f_{\mathrm{rast}}(f_{\mathrm{mlp}}(\mathcal S^{(t)}); \mathrm{\Pi}^{\mathrm{src}})\rVert_2$
		\State $\mathcal S^{(t+1)} =  \mathcal S^{(t)} + \gamma(t) \cdot G_\theta (\mathcal S^{(t)}, \nabla_{\mathcal S^{(t)}} ; t)$
		\State $\text{loss} = \mathcal{L}(f_{\mathrm{rast}}(f_{\mathrm{mlp}}(\mathcal S^{(t + 1)}); \mathrm{\Pi}), \rmI)$
		\State $\text{loss.backward}()$
		\State $\text{update}  \ G_\theta \ \text{and} \  f_{\mathrm{mlp}}$
		\EndFor
		\EndWhile
	\end{algorithmic}
\end{algorithm}

\subsection{\methodname{} Implementation Details}
\label{sec: g3r_supp_implement}

\paragraph{Scene Representation:}
We develop our model based on the 3DGS implementation\footnote{\url{https://github.com/wanmeihuali/taichi_3d_gaussian_splatting}}
~\cite{hu2019taichi}. %
We disable spherical harmonics in our model for simplicity and efficiency following~\cite{luiten2023dynamic}. Moreover, we empiricially find the performance drops are minor when disabling spherical harmonics, as also observed in 3DGS~\cite{3dgs}.
The dimension $C$ of the feature vector $h_i \in \mathbb R^{C}$ is set to 46, with 32 for the latent feature and the remaining 14 for Gaussian attributes including position ($\mathbb R^3$), scale ($\mathbb R^3$), orientation ($\mathbb R^4$), color ($\mathbb R^3$), and opacity ($\mathbb R^1$).

\paragraph{Reconstruction Network (G3RNet):}
We use two generalizable networks with the same architecture for the static background and dynamic scene. We borrow the encoder-decoder UNet architecture from SparseResUNet in \texttt{torchsparse}~\cite{tangandyang2023torchsparse} and do not tune the architecture.
The 3D neural Gaussians and gradients are concatenated as the input of G3R-Net. The timestep positional encodings are concatenated with points’ features output from the last encoder layer and fed to the decoder.
For the background reconstruction network, we use a 2D CNN with 2 residual blocks, without downsampling or upsampling.
For the transformation MLP network $f_\mathrm{mlp}$ that converts the 3D neural Gaussians to a set of explicit 3D Gaussians, we adopt one linear layer with a \texttt{tanh} activation.
The output is combined with a learning rate decay factor $\gamma(t)$ to ensure gradual updates.
The input raw gradient values are normalized for each channel by dividing them by the maximal absolute value in that channel.

\paragraph{Training and Inference:}
During training, we subsample 800k points in total for the static background and dynamic actors to fit into GPU memory. During inference, we subsample 3 million points for higher photorealism. %
To model the sky, we use a sphere image with a fixed radius (\ie, 2048 meters to the center of the ego vehicle at the last frame). As most parts of the sky scene are not visible in the camera, we further crop the top and bottom part of the sphere to only keep the region
between $30^\circ {N}$ and $15^\circ {S}$
to reduce the memory usage.
We initialize the sky points with a resolution of $512\times2048$ during training, while using $1024\times 4096$ during inference.
We select closest 10 source and target frames to train the model.
To produce the camera simulation results in \ref{sec:supp_controllable_sim} and \textbf{supplementary\_video.mp4}, we use all source images. $\lambda_{\mathrm{lpips}}$ and $\lambda_{\mathrm{reg}}$ are both set to $0.01$.
We train our model on  front-facing camera and filter actors/points that are not visible in the field of view.
For multi-camera simulation, we finetune the model on all cameras for 100 iterations.
To further speed up the reconstruction while slightly reducing the photorealism,
we also introduce \methodname{}~(turbo), where we reduce the number of static/dynamic points to 1.5 million, the sky resolution to $512\times2048$, and the number of reconstruction steps to 12.

\paragraph{Additional details for BlendedMVS:}
We initialize 3D Gaussian points by sampling on the surface of provided mesh. We use the high-resolution ($1536 \times 2048$) images. %
During training we take 25 input source images, and 25 as novel views.
There are no dynamic actors in BlendedMVS, so we only model the static background in G3R.
We also do not model a sky-region, as distant regions not covered by the mesh are masked out in the input images.
During training, we subsample 1.5 million points, while during inference we subsample 3.5 million points.
The \textit{turbo} version for BlendedMVS is with 2.5 million points and 24 update steps. However, in each update step, half of the source images are subsampled (each scene has an average of 381 images). %

\section{Implementation Details for Baselines}
\label{sec:baselines}

We now review generalizable reconstruction baseline methods and per-scene optimization methods we compare against.
Unless stated otherwise, we train all generalizable approaches using the same training data as G3R and optimize 3D representations of validation scenes individually with the same source frames for per-scene optimization approaches.

\subsection{MVSNeRF}
MVSNeRF~\cite{chen2021mvsnerf} is a generalizable radiance field reconstruction method that employs a deep neural network to process a few nearby input views and generate the radiance fields representation.
Specifically, it builds a plane-swept 3D cost volume by warping 2D image features (inferred by a 2D CNN) from input views.
Then it leverages a 3D CNN to reconstruct a neural scene volume, encoding both local scene geometry and appearance information.
This 3D neural scene volume is decoded with a
multi-layer perceptron (MLP) to infer density and radiance at arbitrary continuous locations using tri-linearly interpolated neural features inside the scene volume.
Following the original paper, to enhance the rendering realism and leverage more input frames, we fine-tune the neural scene volume along with the MLP decoder for one epoch (around 30 minuites).
We run the official repository\footnote{\url{https://github.com/apchenstu/mvsnerf}} on PandaSet in our experiments.
To handle unbounded driving scenes, we set the maximum rendering range to be $300$ meters for each frame and sample $128$ points for each ray during volume rendering.

\subsection{ENeRF}
ENeRF~\cite{lin2022efficient} constructs a sequential cost volume to predict the approximate geometry and conducts efficient depth-guided sampling. To meet the requirements of the CNN used in ENeRF, we crop the image to $1920 \times 1056$ on PandaSet so that the image dimensions are divisible by 32.
Due to GPU memory contraints, we downscale the images $2\times$ on PandaSet and BlendedMVS during training, but
during inference we use the original full resolution.
We train two models from scratch on PandaSet and BlendedMVS training scenes for 300 epochs using the official repository\footnote{\url{https://github.com/zju3dv/ENeRF}}. We adopt the expoential learning rate decay schedule with \texttt{gamma=0.5} and \texttt{decay\_epochs=50}
During training, we select 4 source images with the closest viewpoints to
each target view.
We choose 2 source images for PandaSet and 4 for BlendedMVS during inference as it empirically produces the best performance.
When taking more source images (\ie 5), ENeRF produces more blurry results (-0.73 drop in PSNR on PandaSet) due to geometry inaccuracy and dynamics.

\subsection{GNT}
GNT~\cite{wang2022attention} samples points along each target ray and predicts the pixel color by learning the aggregation of view-wise features from the epipolar lines using transformers.
We adopt the official repository\footnote{\url{https://github.com/VITA-Group/GNT}} and use \texttt{gnt\_realestate} config to train the models on PandaSet and BlendedMVS. Specifically, we use the original image resolution and train each model for 250k and adjust the batch size to fit within 24GB GPU memory. We choose 2 source views on PandaSet and 10 for BlendedMVS to increase the coverage.
When taking more source images (\ie 5), GNT produces more blurry results (-1.98 drop in PSNR on PandaSet) due to geometry inaccuracy and dynamics.
During inference, we sample 192 points per pixel as suggested by the official guidelines.

\subsection{PixelSplat}
Concurrent work PixelSplat~\cite{charatan2023pixelsplat} predicts 3D Gaussians with
a 2-view epipolar transformer to extract features and then predict the depth distribution and pixel-aligned Gaussians. We adopt the official repository\footnote{\url{https://github.com/dcharatan/pixelsplat}} and use $2\times$ A6000 (48GB) to train the models.
Due to the GPU memory constraint, we downscale the image resolution to $360\times640$ for PandaSet and $384 \times 512$ for BlendedMVS.
We note that the original work uses an
80GB A100 for training and handles $256 \times 256$ resolution.
We use \texttt{re10k} config and train each model for 100k iterations with \texttt{batch\_size=1}.

PixelSplat cannot handle large view changes and produces
rendering results with significant visual artifacts due to inaccurate geometry estimation (\eg, blurry appearance) especiallly on BlendedMVS.
To address this issue, we enhance
PixelSplat, named PixelSplat++, to leverage the 3D scaffold to reduce ambiguity and take all available source images for good coverage.
Specifically,  we first initialize a unified 3D Gaussian representation, unproject DINO~\cite{oquab2023dinov2} image features to 3D points and then use a shared decoder to predict the 3D Gaussian residues.
Similar to G3R, we randomly select one target view, and then choose 10 nearest source views and additional 9 nearest target views during training. We use both the source and target views to supervise the shared decoder and adopt L2 and LPIPS losses.
Compared to PixelSplat, PixelSplat++ takes all source images (original resolution: $1536 \times 2048$) as inputs and predicts a higher-quality 3D representation, achieving a signficiant performance boost at novel views.

\subsection{Instant-NGP}

Instant-NGP~\cite{mueller2022instant} introduces efficient hash encoding, accelerated ray sampling, and fully fused MLPs to neural volumetric rendering.
In our experiments, we use the official repository\footnote{\url{https://github.com/NVlabs/instant-ngp}} and normalize the scenes to occupy the unit cube and set \texttt{aabb\_scale} as 32 for PandaSet and 8 for BlendedMVS to handle the background regions (\eg, far-away buildings and sky) outside the unit cube.
We further enhance Instant-NGP with depth supervision for better performance.
Sepcifically, we aggregate the recorded LiDAR data and create a surfel triangle representation based on estimated per-point normals. Then we render a pseudo-ground-truth depth image at each camera training viewpoint, which is used for depth supervision.
The models are trained for 20k iterations on PandaSet scenes and 100k on BlendedMVS, and converge on the training views.

\subsection{3DGS}
The vanilla version of 3D Gaussian Splatting (3DGS) does not support dynamic scenes or unbounded regions such as the sky.
We therefore employ the same extended version with decomposed foreground, background, and distant regions as in \methodname.
The 3DGS baseline used in this study can be considered as replacing G3RNet during inference with a fixed Stochastic Gradient Descent (SGD) update.
More specifically, we utilize the Adam optimizer with a learning rate of $0.1$ and apply learning rate decay by a factor of 0.5 at iterations 200, 300, 400, and 450. The training process is conducted for a total of 500 iterations. Training for longer iteration does not further improve the performance on the validation views.
It is worth noting that, in each iteration, we aggregate gradients from all source images, which contrasts with other approaches that typically use a single source image per iteration.
Aggregating gradients from all source frames
improves performance and enables more stable training.
We employ the same number of Gaussian points in 3DGS optimiaztion as in \methodname{} inference stage. %
Note that we remove adaptive density control in our experiments as it does not help 3DGS much in test views when it has dense initialization, unless we allow it to grow significantly more points (PSNR+0.58 with 50\% more points (5.3M) in BlendedMVS), at the cost of increased resources.
We also note that enhancing 3DGS with neural Gaussians leads to better results (+0.38 PSNR) and faster early convergence.

\subsection{Efficiency Comparison}
Tab.~\ref{tab:efficiency}, reports the model capacity and training efficiency of baselines and G3R.
G3R's capacity and  efficiency is on par with generalizable methods.
\begin{table}[htbp!]
	\caption{\textbf{Model capacity and training effiency of generalizable approaches.}}
	\renewcommand{\arraystretch}{1.1}
	\centering
	\begin{tabular}{l|cccc}
		\hline
		Method & Train time & Train mem & Recon mem & \#param \\ \hline
		ENeRF           &  108 hours &	24GB & 10GB & $4.3 \times 10^5$  \\
		GNT             &  49 hours &	23GB &	21GB &	$8.8 \times 10^5$        \\
		PixelSplat &  110 hours	& 48GB &	11GB & $1.3 \times 10^8$   \\
		G3R             & 60 hours           & 20GB      & 24GB        & $2.6 \times 10^7$     \\
		\hline
	\end{tabular}
	\label{tab:efficiency}
\end{table}

\section{Experiment Details}
\label{sec:exp_setup}
\subsection{Experiment Setup}
We conduct experiments on two public datasets with large real-world scenes: PandaSet~\cite{xiao2021pandaset} and BlendedMVS~\cite{yao2019blendedmvs}.
PandaSet contains 103 urban driving scenes, each with 6 HD ($1920\times1080$) cameras and LiDAR sweeps.
We select  7 diverse scenes (\texttt{001, 030, 040, 080, 090, 110, 120}) for testing and the remaining are used
for training.
We consider the front camera only for all baselines and G3R for quantitative evaluation experiments.
BlendedMVS-large is a collection of 29 real-world scenes captured by a drone. We use high-resolution ($1538\times 2048$) images in our experiments.
The list of \textit{large scenes} are based on github split\footnote{\url{https://github.com/kwea123/BlendedMVS_scenes/}}
We select 4 scenes for testing (\texttt{58eaf1513353456af3a1682a,
	5b69cc0cb44b61786eb959bf,5bf18642c50e6f7f8bdbd492, 5af02e904c8216-\\ 544b4ab5a2}), each containing $68$ to $836$ images (381 on average).
Unless stated otherwise, for both datasets, we use every other frame as source and the remaining for test.
We use all available images in the supplementary camera simulation demonstrations for novel scene manipulations
such as sensor shifts and actor editing in Sec 4.2 and Sec~\ref{sec:supp_controllable_sim}.

\subsection{Metrics}
We report peak signal-to-noise ratio (PSNR), structural similarity (SSIM)~\cite{wang2004image} and perceptual similarity (LPIPS)~\cite{zhang2018unreasonable} to evaluate the photorealism of novel view synthesis.
To measure the efficiency of different approaches, we also report the reconstruction time and rendering FPS using a single RTX 3090. We note that the generalizable approaches (\eg, ENeRF, GNT, PixelSplat) usually need to reconstruct the scene again with different source images when rendering at new target views
We report the reconstrucion time for one feed-forward prediction.
For MVSNeRF, we report the prediction + finetuning time in Tab. 1.
In contrast, the per-scene optimization methods, PixelSplat++, and G3R obtain a unified representation that takes all input views into account.

\subsection{Evaluation on BlendedMVS}
\begin{figure}[t]
	\centering
	\includegraphics[width=0.99\textwidth]{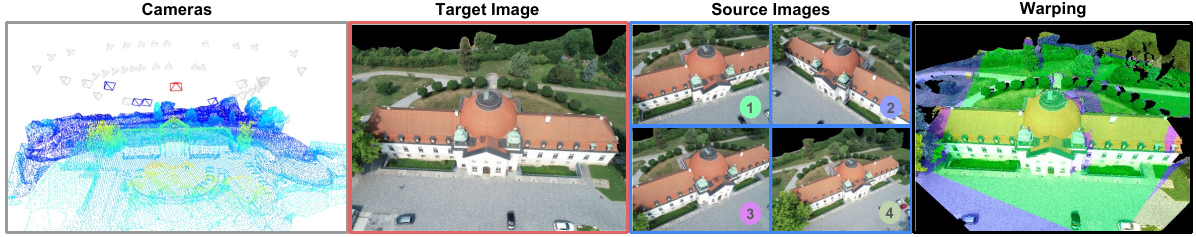}
	\vspace{-0.1in}
	\caption{\textbf{Large view changes on BlendedMVS.} We highlight the \textcolor{red}{target view} in red and 4 closest \textcolor{blue}{source views} in blue.
		The distance and view-orientation changes between the source views and the target view are large.
		The image warping (rightmost column, colored by image source index, missing regions in black) shows that limited source views cannot get full coverage to synthesize the target view.}
	\vspace{-0.15in}
	\label{fig:blendedmvs}
\end{figure}

BlendedMVS has more challenging novel views, as the distance between two nearby views can be large as shown in \cref{fig:blendedmvs}.
We note that there is no explicit interpolation/extrapolation split for BlendedMVS as the multi-pass drone trajectories are not available.

\subsection{Comparison with Generalizable Baselines}
We note that generalizable baselines including ENeRF, GNT and PixelSplat can access all source images but cannot take all images at once due to their limitations. In our experiments, we run baselines in PandaSet for each test frame using 2 closest source images.
When taking more source images (\ie 5), warping-based methods such as ENeRF and GNT produce more blurry results (-0.73/-1.98 PSNR) due to geometry inaccuracy and dynamics.
PixelSplat cannot take more than 2 views due to the memory constrains (48GB) as it predicts pixel-aligned Gaussians and the memory increases linearly with the number of input views.
PixelSplat++ takes all source images as input but it is still worse than G3R as the single-step prediction has limited capacity.

\section{Additional Experiments and Analysis}
We provide additional results and analysis for scene reconstruction on PandaSet and BlendedMVS. We then showcase more camera simulation examples and a generalization study on Waymo Open Dataset (WOD) using G3R.

\subsection{Additional Qualitative Examples}
\label{sec:qualitative_supp}
\begin{figure}[htbp!]
	\centering
	\includegraphics[width=0.99\textwidth]{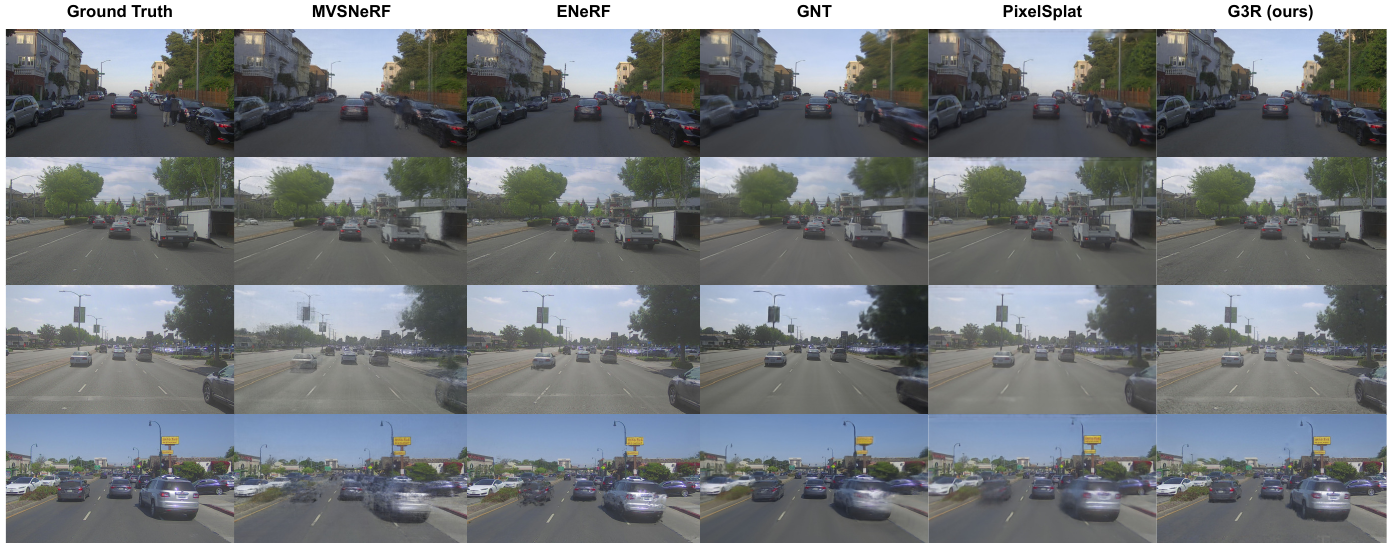}
	\vspace{-0.1in}
	\caption{\textbf{Qualitative comparison to generalizable approaches on PandaSet.}}
	\vspace{-0.05in}
	\label{fig:pandaset_gen_comp}
\end{figure}

\begin{figure}[htbp!]
	\centering
	\includegraphics[width=0.99\textwidth]{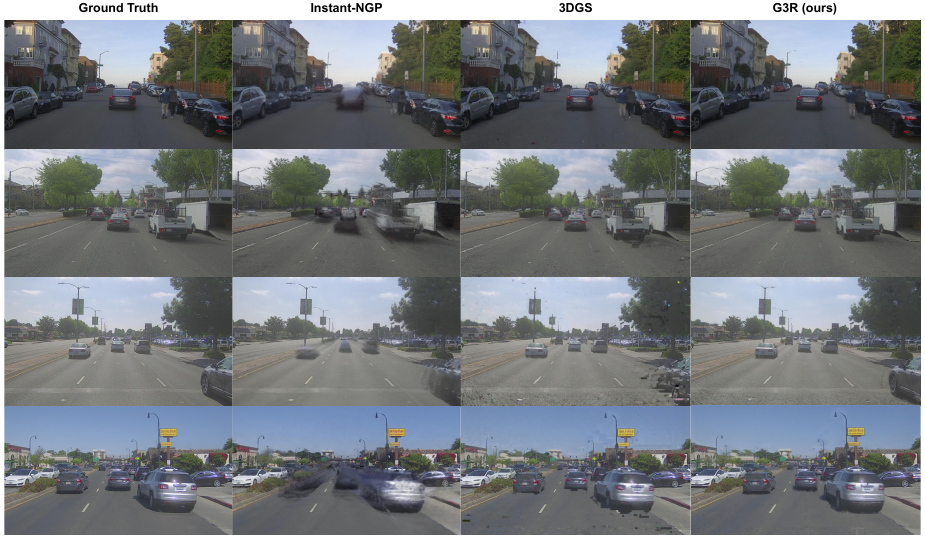}
	\vspace{-0.1in}
	\caption{\textbf{Qualitative comparison to per-scene optimization approaches on PandaSet.}}
	\vspace{-0.05in}
	\label{fig:pandaset_opt_comp}
\end{figure}

We provide additional qualitative comparison with state-of-the-art (SoTA) scene reconstruction approaches on PandaSet. As shown in \cref{fig:pandaset_gen_comp}, compared to G3R, exsiting SoTA generalization approaches suffer from noticeable artifacts such as blurry rendering results, unnatural discontinuities and inaccurate color palette.  In \cref{fig:pandaset_opt_comp}, we further compare G3R with SoTA per-scene optimization approaches. Instant-NGP has severe artifacts on dynamic actors due to lack of dynamics modelling and 3DGS
can produce noticeable artifacts (\eg, black holes) sometimes. In contrast, G3R leads to the most robust rendering results while shortenning the reconstruction times to 2 minutes ($10\times$ speedup).

\begin{figure}[htbp!]
	\centering
	\includegraphics[width=0.99\textwidth]{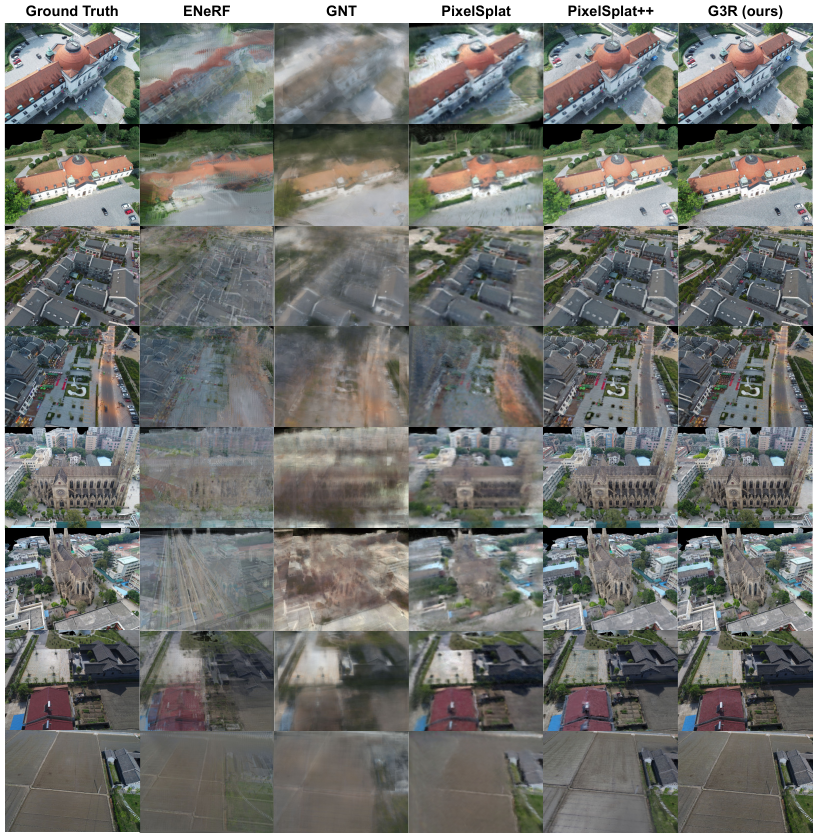}
	\vspace{-0.1in}
	\caption{\textbf{Qualitative comparison to generalizable approaches on BlendedMVS.}}
	\vspace{-0.05in}
	\label{fig:blendedmvs_gen_comp}
\end{figure}

\begin{figure}[htbp!]
	\centering
	\includegraphics[width=0.9\textwidth]{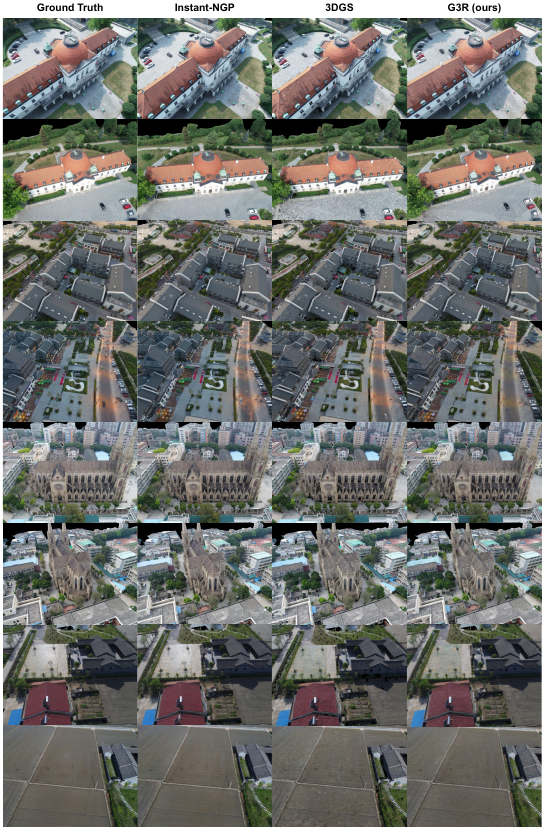}
	\vspace{-0.1in}
	\caption{\textbf{Qualitative comparison to per-scene optimization approaches on BlendedMVS.}}
	\vspace{-0.05in}
	\label{fig:blendedmvs_opt_comp}
\end{figure}

We also present more qualitative comparison with SoTA scene reconstruction approaches on BlendedMVS in \cref{fig:blendedmvs_gen_comp} and \cref{fig:blendedmvs_opt_comp}. As shown in \cref{fig:blendedmvs_gen_comp}, ENeRF, GNT and PixelSplat cannot handle large view changes and produces
rendering results with signficant visual artifacts, including blurry appearance and unnatural discontinuities due to the challenges of estimate high-quality geometry from limited views.
PixelSplat++ achieves a significant performance boost but still produces blurry results compared to G3R due to the chalenge of one-step prediction with limited network capacity.
In \cref{fig:blendedmvs_opt_comp}, we compare G3R with Instant-NGP and 3DGS, and show comparable or better rendering performance with signficiant reconstruction acceleration.

\begin{figure}[htbp!]
	\centering
	\includegraphics[width=0.9\textwidth]{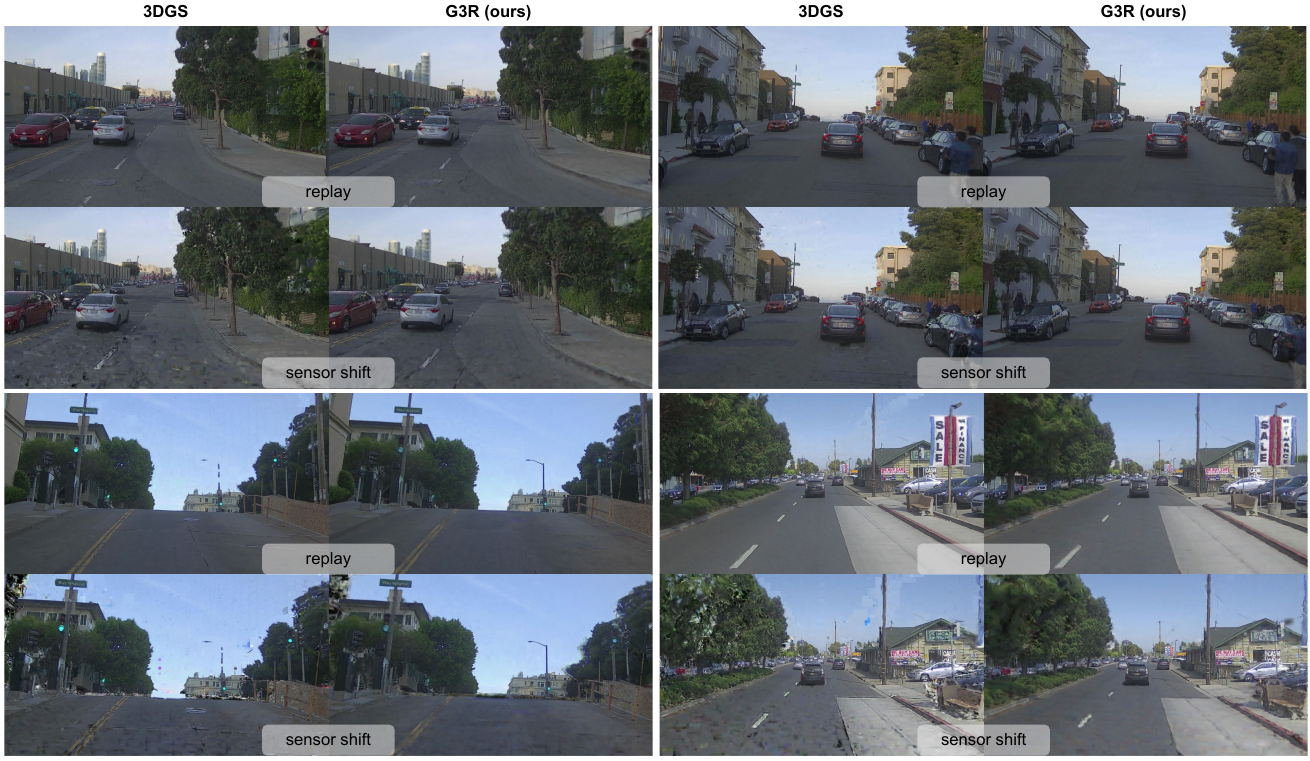}
	\vspace{-0.1in}
	\caption{\textbf{Qualitative comparison of G3R to 3DGS on novel views in PandaSet.}}
	\vspace{-0.05in}
	\label{fig:robust_3dgs}
\end{figure}

\paragraph{Robust 3D Gaussian Prediction}:
To understand why our method achieves superior performance over 3DGS per-scene optimization, we compare the rendering performance at source and novel views. We show a qualitative comparison between 3DGS and \methodname{} where each method gets 20 consecutive frames as input, and then renders the target view several meters forward from the last source view pose (\cref{fig:robust_3dgs}). 
As shown in \cref{tab:opt-comp,tab:extrap-comp}, while 3DGS has sufficient capacity to memorize the source frames, it suffers a significant performance drop (\eg, 1.59 PSNR decrease and 0.054 LPIPS increase) when rendering at novel views. This may be due to the 3DGS-optimized Gaussians having alpha, covariance scales, and orientations that only work well for the source views it's optimized on, resulting in poor underlying geometry~\cite{guedon2023sugar,cheng2024gaussianpro}. 
In contrast, \methodname{} yields more robust Gaussian representations and achieves better rendering performance at novel views on unseen scenes. This is because \methodname{} is trained with novel view supervision across many scenes, which helps regularize the 3D neural Gaussians to generalize rather than merely memorize the source views. As a result, \methodname{} predicts 3D gaussians in a more robust way and produces more realistic rendering performance in both training and extrapolated views.

\begin{table}[t]
	\begin{minipage}{\textwidth}
		\begin{minipage}[c]{0.49\textwidth}
			\caption{\textbf{3DGS overfits to source views while \methodname{} is more robust.}
			}
			\vspace{-0.13in}
			\centering
			\setlength{\tabcolsep}{5pt}
			\resizebox{\textwidth}{!}{
				\begin{tabular}{@{}llll@{}}
					\toprule
					& \multicolumn{1}{l}{PSNR$\uparrow$} & \multicolumn{1}{l}{SSIM$\uparrow$} & \multicolumn{1}{l}{LPIPS$\downarrow$}  \\ \midrule
					\ 3DGS (source views)       &   26.73 & 0.805 & 0.318  \\
					\ Ours (source views)      & 25.94                  & 0.779                   & 0.356                    \\ \midrule
					\ 3DGS (novel views)    &     25.14   &      0.747   &       0.372     \\
					\ Ours (novel views)    & 25.22   & 0.742   & 0.371   \\ \bottomrule
				\end{tabular}
			}
			\vspace{-0.05in}
			\label{tab:opt-comp}
		\end{minipage}
		\hfill
		\begin{minipage}[c]{0.49\textwidth}
			
			\caption{\textbf{Comparison to 3DGS at extrapolated
					views (future 3 frames).}}
			\vspace{-0.13in}
			\centering
			\setlength{\tabcolsep}{5pt}
			\resizebox{\textwidth}{!}{
				\begin{tabular}{@{}llll@{}}
					\toprule
					& PSNR (1st) & PSNR (2nd)   & PSNR (3rd) \\ \midrule
					3DGS~\cite{3dgs}  & 23.96  &  22.43  &   21.58 \\
					Ours &  \textbf{24.13} & \textbf{23.35}  & \textbf{22.82}  \\ \bottomrule
				\end{tabular}
			}
			\vspace{-0.2in}
			\label{tab:extrap-comp}
		\end{minipage}
	\end{minipage}
\end{table}

\subsection{Additional Camera Simulation Examples}
\label{sec:supp_controllable_sim}
\begin{figure}[htbp!]
	\centering
	\includegraphics[width=0.9\textwidth]{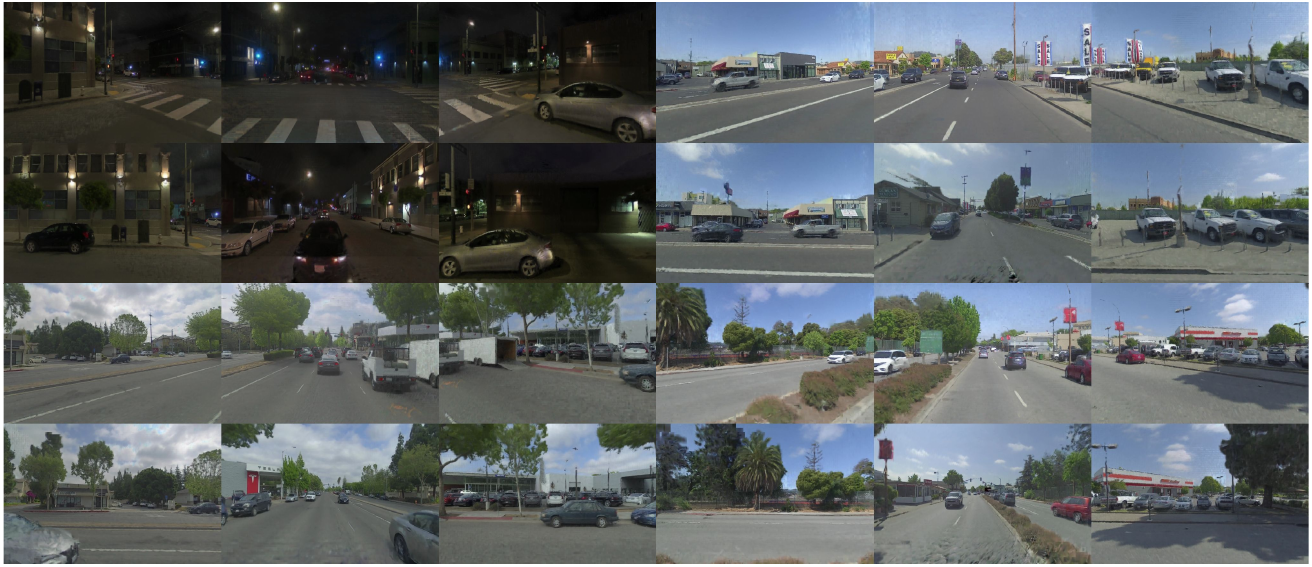}
	\vspace{-0.1in}
	\caption{\textbf{Multi-camera simulation on PandaSet.}}
	\vspace{-0.05in}
	\label{fig:multi_cam_supp}
\end{figure}

\begin{figure}[htbp!]
	\centering
	\includegraphics[width=0.9\textwidth]{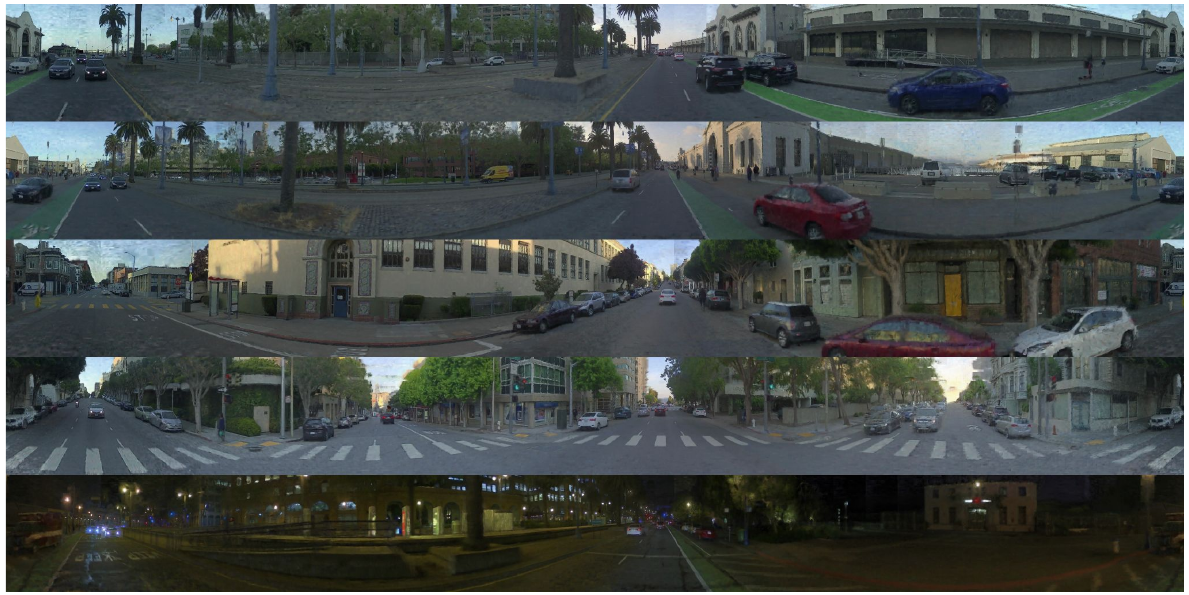}
	\vspace{-0.1in}
	\caption{\textbf{Panorama image simulation on PandaSet.}}
	\vspace{-0.05in}
	\label{fig:pano_supp}
\end{figure}

We now showcase applying G3R for high-fidelity multi-camera simulation for a wide variety of large-scale driving scenes. In \cref{fig:multi_cam_supp} and \cref{fig:pano_supp}, G3R produce consistent and high-fidelity multi-camera or panorama image simulation for diverse scenarios. Please see \cref{sec:limitations} for additional anlaysis on the challenges of multi-camera simulation.

\begin{figure}[htbp!]
	\centering
	\includegraphics[width=0.9\textwidth]{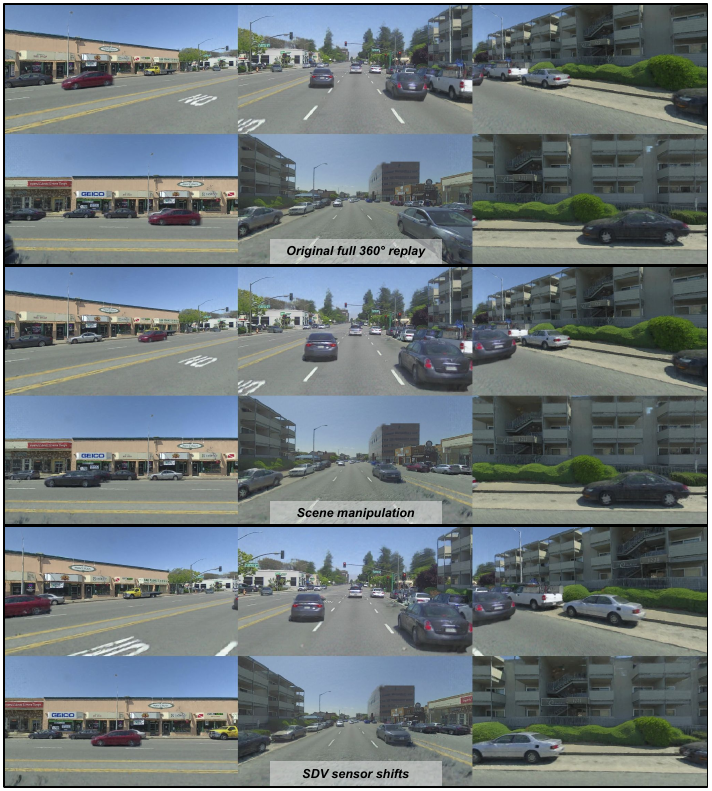}
	\vspace{-0.1in}
	\caption{\textbf{Realistic and controllable multi-camera simulation.}}
	\vspace{-0.05in}
	\label{fig:multi_camera_controllable}
\end{figure}

\begin{figure}[htbp!]
	\centering
	\includegraphics[width=0.9\textwidth]{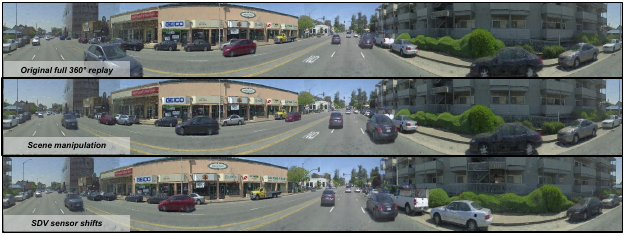}
	\vspace{-0.1in}
	\caption{\textbf{Realistic and controllable panorama image simulation.}}
	\vspace{-0.05in}
	\label{fig:pano_controllable}
\end{figure}

G3R can reconstruct an explicit standalone representation that models the dynamics, which allows us to control, edit and simulate different variations for robotics simulation.
In \cref{fig:multi_camera_controllable} and \cref{fig:pano_controllable}, we show realistic and controllable multi-camera and panorama simulation results by either manipulating the positions of dynamic actors (scene manipulation) or changing
the sensor locations (SDV {camera} sensor shifts).
These results
demonstrate the potential of G3R for scalable self-driving simulation for autonomy validation and training.

\subsection{Additional Generalization Study}
\label{sec:supp_gen_study}
\begin{figure}[htbp!]
	\centering
	\includegraphics[width=0.9\textwidth]{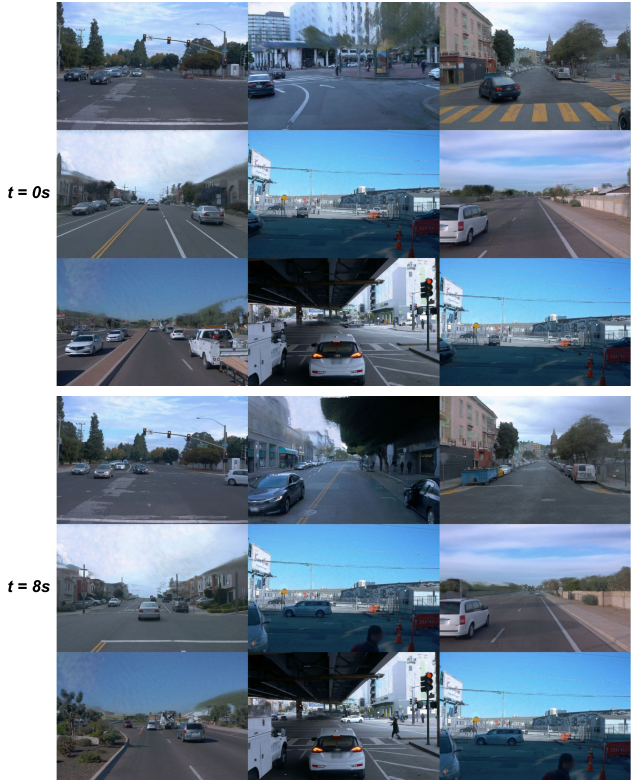}
	\vspace{-0.1in}
	\caption{\textbf{Scene reconstruction on WOD with PandaSet-trained model.}}
	\vspace{-0.05in}
	\label{fig:waymo_supp}
\end{figure}

Finally, we supplement additional results on generalization study across different datasets.
In \cref{fig:waymo_supp}, we directly apply a
pretrained G3R model (on PandaSet) and show it generalizes to new scenes in Waymo Open Dataset~\cite{waymo} (WOD). As shown in \cref{fig:waymo_supp}, G3R can generalize well across datasets with different sensor configurations (placements, sensor type, {appearance} etc) and can reconstruct new scenes in under a few minutes.
This demonstrates
the potential of G3R for
scalable real-world camera simulation.

\subsection{Adaptive Density Control and Robustness Analysis}
\begin{figure}[htbp!]
	\centering
	\includegraphics[width=\columnwidth]{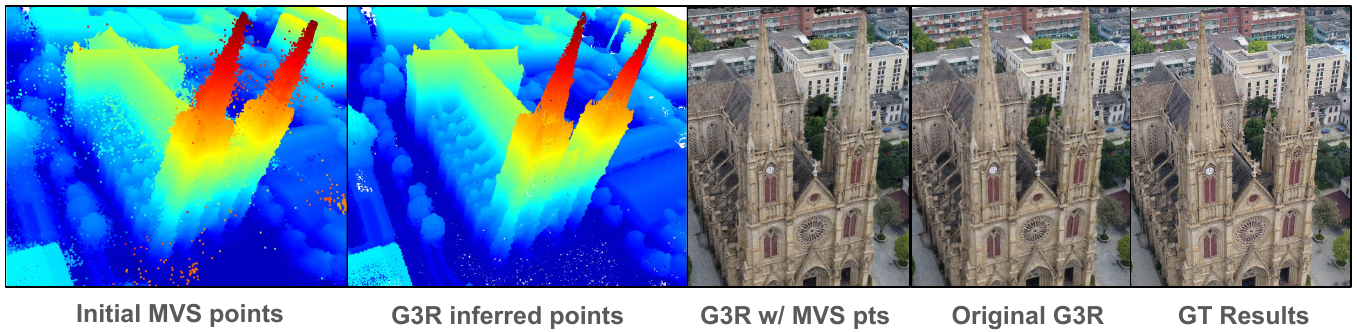}
	\caption{\textbf{G3R is robust to point initialization (zoom-in).}}
	\label{fig:mvs_g3r}
\end{figure}

We experiment with adding density control to G3R and observe enhanced performance.
Specifically, we initialize G3R with 25\% points (0.9M), and grow the points at the 5th step (adding 8 new points around each point and downsample to 3.5M).
The PSNR increases 1.04 compared to no densification, and is 0.42 lower than the original G3R.
While achieving better performance, we notice that G3R has difficulty in handling extremely sparse initialization.
Moreover, we test G3R with dense noisy points from MVS~\cite{wu2024gomvs} (Fig.~\ref{fig:mvs_g3r}) and find G3R is robust to the noisy initialization (only 0.36 PSNR drop).
For robotics applications, dense points from either LiDAR or fast MVS ($\sim$2 min) is typically available.

\section{Limitations and Future Works}
\label{sec:limitations}
\begin{figure}[htbp!]
	\centering
	\includegraphics[width=0.99\textwidth]{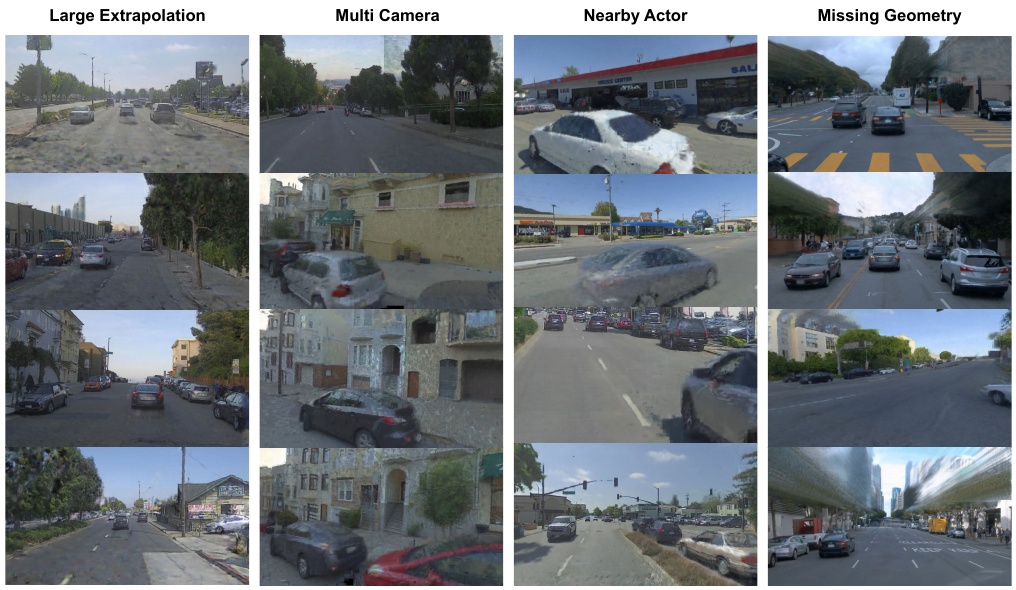}
	\vspace{-0.1in}
	\caption{\textbf{Failure cases of G3R.} }
	\vspace{-0.05in}
	\label{fig:limitations}
\end{figure}

While \methodname{} can reconstruct unseen large scenes efficiently with high photorealism, there are several limitations as shown in \cref{fig:limitations}.
First of all, as shown in \cref{fig:limitations}-leftmost, our approach has artifacts in large extrapolations (\eg, $5\sim10$ meters shift), which may require scene completion and larger scale training to predict novel views with larger differences.
Better surface regularization~\cite{guedon2023sugar,cheng2024gaussianpro} and adversarial training~\cite{roessle2023ganerf,unisim} may mitigate these issues.
Moreover, although \methodname{} shows strong generalizability and robustness thanks to the 3D gradients and recursive updates (\cref{fig:mvs_g3r}), it relies on dense points as initialization and it is an open problem to build effective adaptive density control mechanism for \methodname{} similar to original 3DGS~\cite{3dgs} to prune and grow 3D Gaussians. We notice that the reconstruction quality of \methodname{} degrades on sparse
initialization.

We also do not model non-rigid deformations~\cite{luiten2023dynamic} and emissive lighting for more controllable simulation.
We also notice more artifacts in multi-camera simulation (\cref{fig:limitations}-second-column), primarily due to the different exposure and white balance settings across cameras,
misalignment due to calibration errors,
as well as motion blur and rolling shutter for the side cameras. %
Additionally,
nearby dynamic actors have more artifacts, particularly due to the resolution of the Gaussian points (\cref{fig:limitations}-third-column).
Incorporating multi-resolution or level-of-detail modelling to the neural 3D Gaussians could improve this.
Lastly, there are
artifacts when points are missing for some regions (e.g., the higher part of the building, particularly in the WOD dataset), because these regions are not scanned by the LiDAR and are thus modeled as part of the sky (\cref{fig:limitations}-rightmost).
SfM and MVS points can be added to mitigate this problem~\cite{yan2024street}.

\section{Broader Impact}
\label{sec:broader_impact}

G3R provides a scalable and efficient way to reconstruct large-scale real-world scenes for high-quality and real-time rendering.
Its ability to generate controllable camera simulation videos (\eg, scene manipuation and sensor shifts) can potentially improve the robustness and safety of robotic systems for real-world environments or can be used to build immersive experience in VR/AR applications.
\end{document}